\documentclass[runningheads]{llncs}
\usepackage{graphicx}

\usepackage{tikz}
\usepackage{wrapfig}
\usepackage{comment}
\usepackage{amsmath,amssymb} 
\usepackage{color}
\usepackage{adjustbox}
\usepackage{booktabs}
\usepackage{multirow}
\usepackage{bbding}
\usepackage{pifont}
\usepackage{wasysym}
\usepackage{amssymb}
\makeatletter
  \newcommand\figcaption{\def\@captype{figure}\caption}
  \newcommand\tabcaption{\def\@captype{table}\caption}
\makeatother

\newcommand\blfootnote[1]{%
  \begingroup
  \renewcommand\thefootnote{}\footnote{#1}%
  \addtocounter{footnote}{-1}%
  \endgroup
}

\usepackage[accsupp]{axessibility}  

\RequirePackage{makecell}

\begin{document}
\pagestyle{headings}
\mainmatter

\title{Tip-Adapter: Training-free Adaption of CLIP \\for Few-shot Classification} 


\titlerunning{Tip-Adapter: Training-free Adaption of CLIP}
%
\author{Renrui Zhang\inst{*1,2} \and
Wei Zhang\inst{*1} \and
Rongyao Fang\inst{2} \and
Peng Gao\inst{\dagger1} \and
Kunchang Li\inst{1} \and
Jifeng Dai\inst{3} \and
Yu Qiao\inst{1} \and
Hongsheng Li\inst{2,4}}
\authorrunning{R. Zhang et al.}
%
\institute{Shanghai AI Laboratory \and
The Chinese University of Hong Kong \and
SenseTime Research \and
Centre for Perceptual and Interactive Intelligence (CPII)\\
\email\{zhangrenrui, gaopeng, qiaoyu\}@pjlab.org.cn,\ \ hsli@ee.cuhk.edu.hk}
\maketitle
\begin{abstract}
Contrastive Vision-Language Pre-training, known as CLIP, has provided a new paradigm for learning visual representations using large-scale image-text pairs. It shows impressive performance on downstream tasks by zero-shot knowledge transfer. To further enhance CLIP's adaption capability, existing methods proposed to fine-tune additional learnable modules, which significantly improves the few-shot performance but introduces extra training time and computational resources. In this paper, we propose a \textbf{T}raining-free adaption method for CL\textbf{IP} to conduct few-shot classification, termed as \textbf{Tip-Adapter}, which not only inherits the training-free advantage of zero-shot CLIP but also performs comparably to those training-required approaches. Tip-Adapter constructs the adapter via a key-value cache model from the few-shot training set, and updates the prior knowledge encoded in CLIP by feature retrieval. On top of that, the performance of Tip-Adapter can be further boosted to be state-of-the-art on ImageNet by fine-tuning the cache model for 10$\times$ fewer epochs than existing methods, which is both effective and efficient. We conduct extensive experiments of few-shot classification on 11 datasets to demonstrate the superiority of our proposed methods. Code is released at \url{https://github.com/gaopengcuhk/Tip-Adapter}.

\end{abstract}
\blfootnote{$^*$ Indicates equal contributions, $\dagger$ Indicates corresponding author}

\section{Introduction}

Vision and language are two modalities for humans to perceive the surrounding world and perform diverse interactions with the environment. 
The accuracy of vision tasks, such as classification~\cite{krizhevsky2012imagenet,he2016deep,howard2017mobilenets,dosovitskiy2021vit,mao2021dual,gao2022convmae,zhao2022towards}, detection~\cite{ren2015faster,carion2020end,zheng2020end,xu2021layer,zhang2022monodetr,Cui_2021_ICCV} and 3D understanding~\cite{qi2017pointnet,zhang2022pointclip,xu2021poem,zhang2022point} has been boosted significantly thanks to better neural architecture designs~\cite{he2016deep,vaswani2017attention} and delicately designed frameworks~\cite{ren2015faster,lin2017focal,carion2020end,zhao2022tracking}. 
Language tasks concerning generation and understanding have also been largely improved due to large-scale self-supervised methods, including pre-training by mask prediction~\cite{devlin2018bert} and collected web-scale data~\cite{radford2018improving}. 
As vision and language normally contain complementary information, joint learning of multi-modality representations has been proven to be quite effective on various tasks, such as visual question answering~\cite{antol2015vqa,anderson2018bottom,kim2018bilinear}, image captioning~\cite{you2016image,huang2019attention}, and referring expression~\cite{yu2018mattnet}. 
Different from previous methods that independently learn vision and language representations on separate datasets~\cite{anderson2018bottom,lu2019vilbert,tan2019lxmert}, CLIP~\cite{radford2021learning} proposed to learn transferable visual features from paired natural language supervisions and exerted amazing zero-shot image classification ability. Due to the interplay between language and vision, the encoded visual representations can be used in open-vocabulary recognition without further re-training. 

Many follow-up works have proposed to utilize few-shot data to improve CLIP's adaption capability on downstream tasks. Following the direction of prompt design~\cite{brown2020language,liu2021pre}, CoOp~\cite{zhou2021coop} fine-tuned the pre-trained CLIP via learnable textual tokens and achieved strong performance on few-shot image classification. Recently, 
CLIP-Adapter~\cite{gao2021clip} introduced to equip CLIP with a parametric feature adapter, which generates adapted features and combines them with the original CLIP-encoded features via a residual connection. It demonstrated promising performance for few-shot classification without utilizing prompt designs. Although CoOp~\cite{zhou2021coop} and CLIP-Adapter~\cite{gao2021clip} have shown powerful capabilities on few-shot classification benchmarks, in comparison with Zero-shot CLIP~\cite{radford2021learning} and Linear-probe CLIP~\cite{radford2021learning}, they require more computational resources to fine-tune the newly introduced learnable parameters.
Thus, we ask the following question: {can we achieve the best of both worlds, which not only takes the advantage of CLIP's training-free property for zero-shot classification but also enjoys the strong performance of training-required methods
for few-shot classification?}


\begin{table}[t]
    \label{tab1}
	\centering
	\caption{Comparison of classification accuracy (\%) and time efficiency for different methods on 16-shot ImageNet~\cite{deng2009imagenet}, where our proposed Tip-Adapter and Tip-Adapter-F achieve superior accuracy-efficiency trade-off. All experiments are tested with batch size 32 on a single NVIDIA GeForce RTX 3090 GPU. The column in blue records the performance gain relative to Zero-shot CLIP.}
	\adjustbox{width=0.9\linewidth}{
	\begin{tabular}{lccccccc}
	\toprule
		Models &\makecell*[c]{\ \ Training} \ \  &\makecell*[c]{Epochs} & \makecell*[c]{Time} &  Accuracy &\makecell*[c]{Gain} &\ Infer. Speed \ \ &GPU Mem.
		\\ \midrule
		Zero-shot CLIP~\cite{radford2021learning} &Free &0 &0 &60.33 &0 &10.22ms &2227MiB\\
		Linear-probe CLIP~\cite{radford2021learning} &Required & - & 13min & 56.13 &\color{blue}{$-4.20$}&- &-\\
		CoOp~\cite{zhou2021coop} &Required& 200 &14h\ 40min & 62.95  &\color{blue}{$+2.62$} &299.64ms &7193MiB\\
		CLIP-Adapter~\cite{gao2021clip} &Required& 200 & 50min & 63.59 &\color{blue}{$+3.26$} &10.59ms &2227MiB\\ \midrule
		Tip-Adapter &\textbf{Free}& \textbf{0} & \textbf{0} & 62.03\ & \color{blue}{$+1.70$} &10.42ms &2227MiB\\
	    Tip-Adapter-F &Required& 20 & 5min & \textbf{65.51} & \color{blue}{$+5.18$} &10.53ms &2227MiB\\
	\bottomrule
	\end{tabular}}
\end{table}



To achieve the goal, we propose a \textbf{T}raining-free adaption method for CL\textbf{IP}, named \textbf{Tip-Adapter}, which appends the weight-frozen CLIP model with a novel non-parametric adapter. Different from existing methods, ours does not require extra training, but designs the adapter as a query-key cache model~\cite{khandelwal2019generalization,orhan2018simple,grave2017unbounded} from the few-shot dataset. Specifically, Tip-Adapter extracts visual features of few-shot images by CLIP's visual encoder and transforms their corresponding labels into one-hot encodings. Then, a cache model containing few-shot visual features and one-hot labels is created, which are viewed as paired keys and values.

By the cache model, the training-free construction of Tip-Adapter exhibits great efficiency compared to traditional fine-tuning via Stochastic Gradient Descent (SGD)~\cite{kingma2014adam,DBLP:conf/iclr/LoshchilovH19}. During inference, the test image first calculates its feature similarities with cached keys, and then aggregates cached values to form the adapter's prediction, which can be regarded as retrieving the few-shot knowledge from the cache model. After that, the adapter's prediction is combined with the original CLIP's prediction by a residual connection~\cite{he2016deep}. In this way, Tip-Adapter simultaneously exploits knowledge from both pre-trained CLIP and the few-shot training dataset.
Surprisingly, Tip-Adapter without training could perform comparably to the fine-tuned CoOp and CLIP-Adapter. 
Furthermore, if we unfreeze the cached keys as learnable parameters and further fine-tune them, Tip-Adapter's performance could be significantly boosted with just a few training epochs. We term this fine-tuned version as \textbf{Tip-Adapter-F}, which only requires 20 epochs on ImageNet~\cite{deng2009imagenet} to be state-of-the-art compared with 200 epochs adopted by CoOp and CLIP-Adapter. In Table~\ref{tab1}, we list the comparison between all existing methods of their performance, training time and inference speed for 16-shot classification on ImageNet, which indicates great accuracy-efficiency trade-off of our methods. 

The contributions of our paper are summarised below:
\begin{enumerate}
    \item We propose Tip-Adapter, a training-free adaption method for CLIP, which discards the conventional SGD-based training by directly setting the adapter with a cache model.
    
    \item Unfreezing the keys of cache model as learnable parameters, the fine-tuned Tip-Adapter, named Tip-Adapter-F, achieves state-of-the-art performance with super-fast convergence on ImageNet.
    
    \item We evaluate Tip-Adapter and Tip-Adapter-F on 11 widely-adopted datasets for few-shot classification and conduct extensive ablation studies to demonstrate their characteristics.
\end{enumerate}

\section{Related Work}

\paragraph{\textbf{Data-efficient Transfer Learning.}}
The capability of deep neural networks is revealed with the assistance of large-scale and high-quality datasets~\cite{krizhevsky2012imagenet}. However, collecting such data is challenging and expensive due to long-tail distributions, noisy annotations and the increasing labeling cost. Thus, transfer learning is proposed to alleviate this issue, which has become a popular research field. Supervised pre-training on image classification~\cite{deng2009imagenet} have been widely adopted as a default basis for fine-tuning on downstream tasks (e.g. detection~\cite{ren2015faster} and segmentation~\cite{he2017mask}).
Self-supervised learning, such as MoCo~\cite{he2020momentum} and BYOL~\cite{grill2020bootstrap}, further discards the need of supervised signals and builds a contrastive pretext task for robust feature learning. Recently, CLIP~\cite{radford2021learning}, DeCLIP~\cite{li2021supervision} and ALIGN~\cite{jia2021scaling} demonstrate that learning from simple contrastive vision-language pairs obtains promising transferable features for zero-shot recognition over diverse datasets. On top of that, CoOp~\cite{zhou2021coop}, CLIP-Adapter~\cite{gao2021clip} and WiSE-FT~\cite{wortsman2021robust} significantly improve the CLIP with limited training data by freezing the pre-trained weights and training additive learnable modules. In contrast, our proposed Tip-Adapter aims at directly infusing few-shot supervisions into the pre-trained CLIP model in a training-free manner.
By this, the construction of Tip-Adapter is much more efficient for both time and memory, which only requires calculating the features of few-shot training set once and then caches them.


\paragraph{\textbf{Cache Model.}}
A cache model stores features of training images and their labels as a key-value database. During inference, the feature encoded from a test sample is treated as query to aggregate information from the cache model by similarity-based retrieval~\cite{vaswani2017attention}. The whole process is non-parametric~\cite{kossen2021self} and involves no parameter update. The cache model has been equipped on various models to boost the performance for vision or language models, including kNN-LMs~\cite{khandelwal2019generalization}, Unbounded Cache~\cite{grave2017unbounded}, Matching Network~\cite{vinyals2016matching} and others~\cite{merity2016pointer,snell2017prototypical}. Although simple cache model~\cite{orhan2018simple} has shown promising results, the huge storage budget for training data is unaffordable to many applications. To reduce such cost, approximate kNN with highly-optimized similarity search system~\cite{johnson2019billion} is proposed, which however is slow and error-prone. Different from previous setup with pure vision or language caches, we construct a blended vision-language cache model by CLIP's contrastive multi-modality pre-training. Importantly, thanks to our few-shot setting with limited training samples, the total cache size is small and the retrieval can be efficiently calculated by two cascaded matrix multiplications. Moreover, the cache model in Tip-Adapter can be learnable and dynamically updated by Stochastic Gradient Descent (SGD), which further improves its performance.

\section{Method}
In Section~\ref{tip}, we first introduce our proposed training-free Tip-Adapter and its fine-tuned variant, Tip-Adapter-F. Then in Section~\ref{cache}, we discuss the relations between our approach and previous methods, such as CLIP-Adapter and cache-based networks.

\subsection{Training-free Adaption of CLIP}\label{tip}
We propose Tip-Adapter, a training-free adaption method to enhance the few-shot classification performance of CLIP. 
We construct a key-value cache model from the few-shot training set in a non-parametric manner. Surprisingly, with this well-designed cache model, Tip-Adapter without fine-tuning can achieve comparable performance compared to those training-required approaches, including CoOp~\cite{zhou2021coop} and CLIP-Adapter~\cite{gao2021clip}. In addition, if training is allowed, Tip-Adapter-F further surpasses state-of-the-art performance by fine-tuning the cached keys with super-fast convergence.

\paragraph{\textbf{Cache Model Construction.}}
Given the pre-trained CLIP~\cite{radford2021learning} model and a new dataset with $K$-shot $N$-class training samples for few-shot classification, there are $K$ annotated images in each of the $N$ categories, denoted as ${I}_{K}$ with their labels ${L}_{N}$. We aim at creating a key-value cache model as the feature adapter, which contains few-shot knowledge within $N$ classes. For each training image, we utilize the CLIP's pre-trained visual encoder to extract its $C$-dimensional L2 normalized feature, and convert its ground-truth label into an $N$-dimensional one-hot vector. For all $NK$ training samples, we denote their visual features and corresponding label vectors as $\mathbf{F}_{\rm train} \in \mathbb{R}^{NK \times C}$ and $\mathbf{L}_{\rm train} \in \mathbb{R}^{NK \times N}$,
\begin{align}
    &\mathbf{F}_{\rm train} = \mathrm{Visual Encoder}({I}_
    {K}),\\ &\mathbf{L}_{\rm train} = \mathrm{One Hot}({L}_{N}).
\end{align}
For the key-value cache, the CLIP-encoded representations $\mathbf{F}_{\rm train}$ are treated as keys, while the one-hot ground-truth vectors $\mathbf{L}_{\rm train}$ are used as their values. In this way, the key-value cache memorizes all the new knowledge extracted from the few-shot training set, which is for updating the prior knowledge encoded in the pre-trained CLIP.

\begin{figure*}[t!]
  \centering
    \includegraphics[width=1\textwidth]{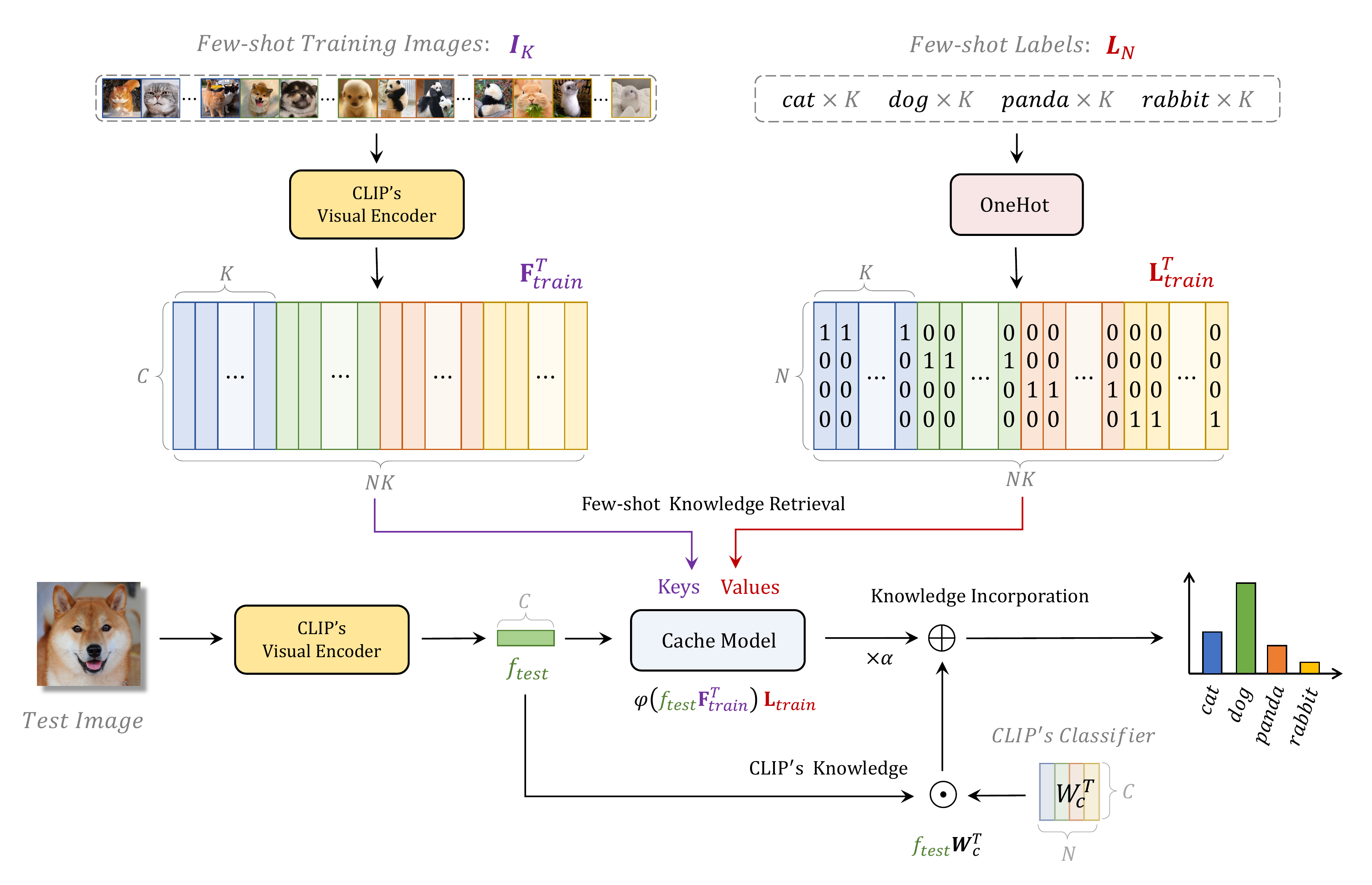}
   \caption{\textbf{The Pipeline of Tip-Adapter.} Given a $K$-shot $N$-class training set, we construct a cache model to adapt CLIP on downstream tasks. It contains few-shot visual features $\mathbf{F}^T_{\rm train}$ encoded by CLIP and their ground-truth labels $\mathbf{L}^T_{\rm train}$ under one-hot encodings. After retrieval from the cache model, the few-shot knowledge is incorporated with CLIP's pre-trained knowledge, achieving the training-free adaption.}
    \label{fig:cache_model}
\end{figure*}

\paragraph{\textbf{Tip-Adapter.}}
After constructing the cache model, the adaption of CLIP can be simply achieved by two matrix-vector multiplications. During inference, the L2 normalized feature $f_\mathrm{test} \in \mathbb{R}^{1 \times C}$ of the test image is first extracted by the CLIP's visual encoder and serves as a query for retrieving from the key-value cache. The affinities between the query and keys can be estimated as
\begin{align}
\label{beta}
      A =  \exp{(-\beta(1 - f_\mathrm{test} \mathbf{F}^T_{\rm train}))},
\end{align}
where $A \in \mathbb{R}^{1 \times NK}$ and $\beta$ stands for a modulating hyper-parameter. Since both query and key features are L2 normalized, the term $f_\mathrm{test} \mathbf{F}^T_{\rm train}$ is equivalent to the cosine similarities between test feature $f_\mathrm{test}$ and all few-shot training features $\mathbf{F}^T_{\rm train}$. 
The exponential function is adopted to convert the similarities into non-negative values with $\beta$ modulating its sharpness. 
Afterwards, the prediction for cache model can be obtained via linear combination of  cached values weighted by the query-key affinities, denoted as $A \mathbf{L}_{\rm train} \in \mathbb{R}^{1 \times N}$.

Besides the few-shot knowledge retrieved from cache model, the prior knowledge of pre-trained CLIP is calculated by $f_\mathrm{test} W_c^T \in \mathbb{R}^{1 \times N}$, where $W_c$ is the weights of CLIP's classifier generated from its pre-trained textual encoder. By blending both predictions via a residual connection, the output logits of the test image by Tip-Adapter are then calculated as
\begin{align}
\label{logits}
    \mathrm{logits} &= \alpha A \mathbf{L}_\mathrm{train} + f_\mathrm{test} W_c^T \notag \\
    &= \alpha \varphi(f_\mathrm{test} \mathbf{F}^T_{\rm train}) \mathbf{L}_\mathrm{train} + f_\mathrm{test} W_c^T,
\end{align}
where $\alpha$ denotes the residual ratio, and we define $\varphi(x) = \exp(-\beta(1 - x))$. Tip-Adapter's prediction therefore contains two terms, the former term adaptively summarizes information from the few-shot training dataset, and the latter term preserves the prior knowledge from the CLIP's classifier $W_c^T$.
The two terms are balanced by the weight $\alpha$. Empirically, $\alpha$ is set to be large if the domain gap between pre-trained and downstream few-shot tasks is large, since more knowledge from the few-shot set is required, and small otherwise.

\paragraph{\textbf{Tip-Adapter with Fine-tuning.}}
Tip-Adapter can greatly boost CLIP by incorporating new knowledge in the few-shot training set.
However, given more shots, Tip-Adapter without training gradually lags behind the training-required CoOp and CLIP-Adapter. To mitigate the gap while preserve the efficiency, we propose Tip-Adapter-F, which treats the keys in the cache model as a good initialization for learnable parameters, and fine-tunes them via SGD. Thanks to the advantageous starting point of cache model, Tip-Adapter-F achieves state-of-the-art performance with only 20-epoch fine-tuning on ImageNet~\cite{deng2009imagenet}, compared with CoOp and CLIP-Adapter's 200-epoch training.

More specifically, we unfreeze the cached keys $\mathbf{F}_\mathrm{train}$, but still freeze the values $\mathbf{L}_\mathrm{train}$ and the two encoders of pre-trained CLIP. The intuition is that updating the keys in the cache model can boost the estimation of affinities, which is able to calculate the cosine similarities between the test and training images more accurately. In contrast, values in the cache model are one-hot encodings representing ground-truth annotations and shall be kept frozen to well memorize the category information.

\subsection{Relations with Previous Models}\label{cache}
    
\begin{minipage}[t]{0.54\linewidth}

\centering
\includegraphics[width=\linewidth]{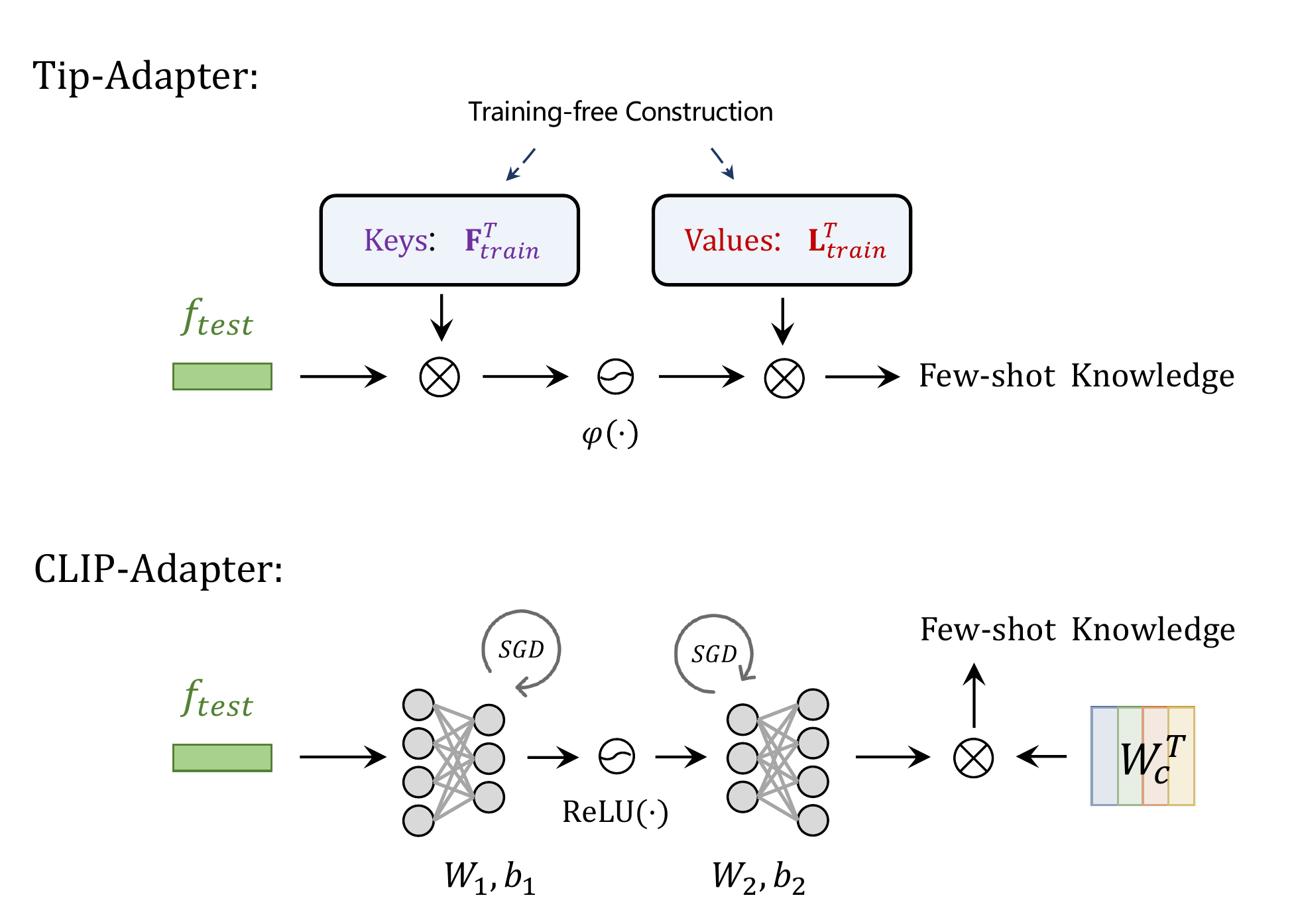}
\figcaption{Comparison of Tip-Adapter and CLIP-Adapter~\cite{gao2021clip} to acquire few-shot knowledge. Tip-Adapter retrieves from the constructed cache model, but CLIP-Adapter encodes the knowledge by the learnbale adapter and obtains it aided by CLIP's classifier $W_c$.}
\label{fig:compare}
\end{minipage}\quad
\begin{minipage}[t]{0.43\linewidth}
\centering
\includegraphics[width=\linewidth]{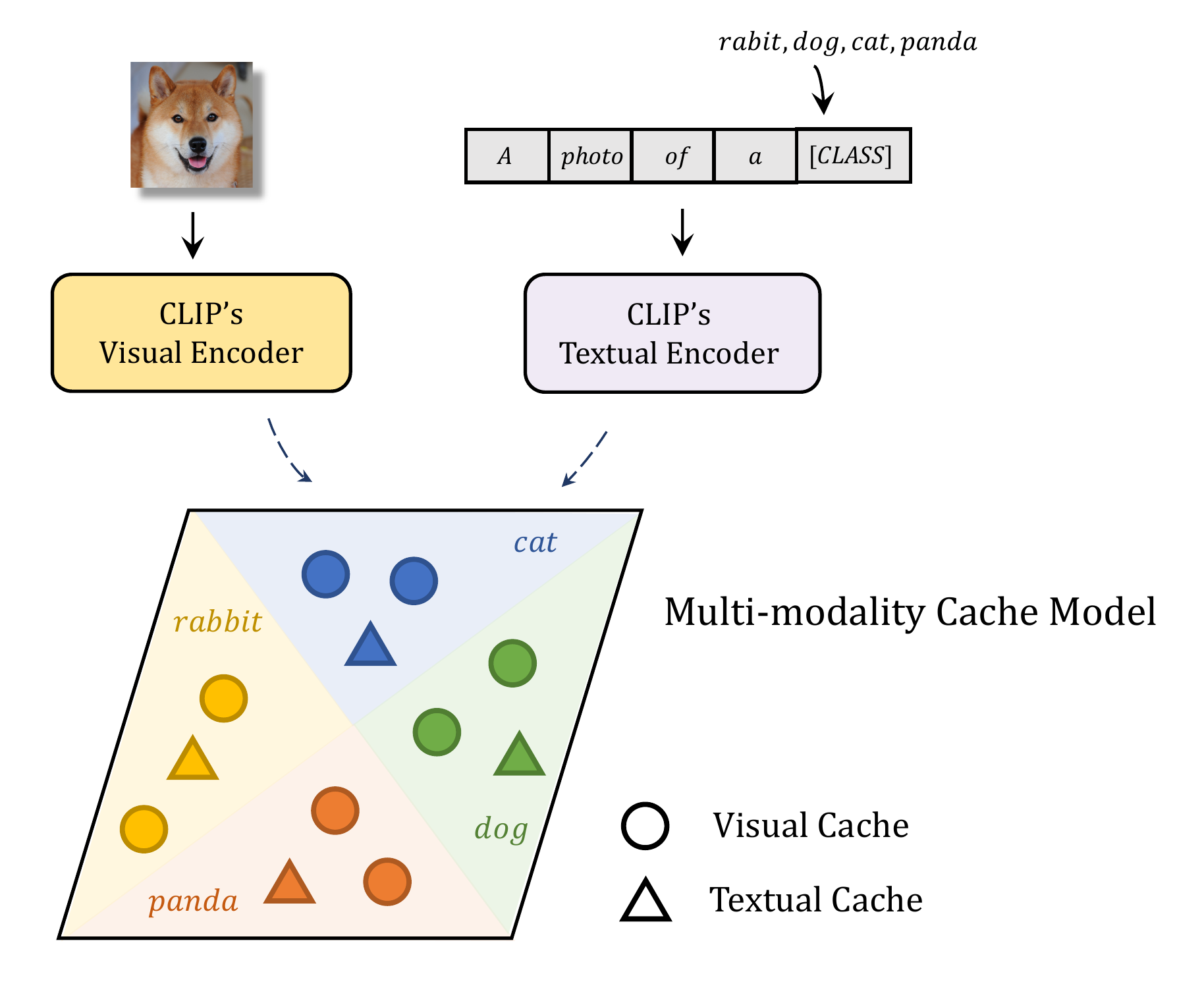}
\figcaption{The multi-modality cache model of Tip-Adapter. Different from previous networks only with visual cache, Tip-Adapter caches both visual and textual knowledge by CLIP's encoders.}
\label{relations}
\end{minipage}

\paragraph{\textbf{Relations with CLIP-Adapter.}}
Following the adapter \cite{houlsby2019parameter} in neural language processing, CLIP-Adapter~\cite{gao2021clip} appends a lightweight two-layer Multi-Layer Perceptron (MLP) to the pre-trained weight-fixed CLIP model and optimizes its parameters via SGD. Specifically, for an input test image, its visual feature $f_{test}$ is first obtained by CLIP's pre-trained visual encoder. Then, the MLP-based adapter with randomly initialized parameters $W_1, b_1, W_2, b_2$, is appended to output the adapted feature,
\begin{align}
\begin{split}
    f_{\rm test}^{a} = \varphi(f_{\rm test} W_1^T + b_1) W_2^T + b_2,
\end{split}
\end{align}
where $\varphi$ denotes the activation function in the MLP. Afterwards, the adapted feature $f^a_{\rm test}$ is linearly combined with the pre-trained CLIP's feature $f_{\rm test}$, and output the final classification logits with a hyper-parameter $\alpha \in [0,1]$,
\begin{align}
\label{v1}
\begin{split}
    \mathrm{logits} = \alpha f^a_{\rm test} W_c^T + f_{\rm test} W_c^T,
\end{split}
\end{align}
where $W_c^T$ is the weights of CLIP's classifier. 
The first terms in both Eqs.~\eqref{logits} and~\eqref{v1} represent the ways of Tip-Adapter and CLIP-Adapter to obtain the few-shot knowledge, respectively. As shown in Figure~\ref{fig:compare}, Tip-Adapter acquires the knowledge by retrieval from the cache model, but CLIP-Adapter first utilizes the learnable adapter to predict the adapted feature and then multiplies it with CLIP's $W_c^T$ to form the final knowledge output.

With further analysis for Eqs.~\eqref{logits} and \eqref{v1}, CLIP-Adapter can be seen as a special form of our proposed Tip-Adapter,
\begin{align}
\label{diffs}
    &W_1 = \mathbf{F}_\mathrm{train}, \ W_2 = \mathbf{L}^T_\mathrm{train} W_c^{-1},
   \ \ b_1 = 0, \ b_2 =0, \\
    &\varphi(x) = \operatorname{exp}(- \beta (1 - x)), \  \text{ where } \  x \in [0, 1].
\end{align}
They have two key differences. Firstly, CLIP-Adapter randomly initializes both keys and values in the cache model as $W_1$ and $W_2$, and learns them via SGD, while Tip-Adapter directly constructs them with cached training features $\mathbf{F}_\mathrm{train}$ and one-hot encodings of the ground-truth labels $\mathbf{L}_\mathrm{train}$, which are non-parametric and training-free. 
Secondly, the bottleneck dimension of Tip-Adapter is equal to $NK$, while, to prevent over-fitting resulted from training, CLIP-Adapter selects a lower-dimensional bottleneck. This indicates that our cache model could better alleviate the over-fitting problem on few-shot datasets, which further releases the fitting power of large-scale pre-trained models.
Thirdly, Tip-Adapter introduces the activation function denoted in Eq.~\eqref{diffs}. As its inputs are the distances in the normalized feature space, it is naturally bounded between 0 and 1. However, for CLIP-Adapter, the common activation function, ReLU$(\cdot)$, is chosen to handle unbounded inputs.
In short, Tip-Adapter obtains a well-performing adapter without training, which is more efficient on few-shot classification.

\paragraph{\textbf{Relations with Cache-based Networks.}}
Acquiring a cache model from few-shot training data has been explored by many previous methods, including Matching Network~\cite{vinyals2016matching}, Prototypical Networks~\cite{snell2017prototypical}, MAML~\cite{finn2017model}, Relation Network~\cite{sung2018learning} and others~\cite{dhillon2019baseline,chen2020new,tian2020rethinking,chen2019closer}. 
Our models differ from them in two points for both specific methods and experimental settings.

Firstly, previous works only constructs the cache of visual features, but Tip-Adapter adopts a multi-modality heterogeneous cache model with both visual and textual cached features extracted by CLIP, as shown in Figure~\ref{relations}. 
In detail, the aforementioned cache model with keys $\mathbf{F}_\mathrm{train}$ and values $\mathbf{L}_\mathrm{train}$ serves as the visual cache, denoted as $\mathbf{F}_\mathrm{vis}$ and $\mathbf{L}_\mathrm{vis}$ here. As CLIP's classifier $W_c$ is calculated from category texts by the textual encoder, $W_c \in \mathbb{R}^{N \times C}$ can be viewed as language features serving as keys $\mathbf{F}_\mathrm{tex}$ for textual cache. The values of textual cache is then denoted by an identity matrix $\mathbf{L}_\mathrm{tex} \in \mathbb{R}^{N \times N}$, since $W_c$ respectively encodes $N$ category knowledge and each of its row vector corresponds to a certain category. From this perspective, Eq.~\eqref{logits} is reformulated as
\begin{align}
\label{logits_cache}
    \mathrm{logits} &= \alpha \varphi(f_\mathrm{test} \mathbf{F}^T_{\rm vis}) \mathbf{L}_\mathrm{vis} + (f_\mathrm{test} \mathbf{F}^T_{\rm tex})\mathbf{L}_\mathrm{tex},
\end{align}
where the two terms represent knowledge retrieval from both visual and textual cached knowledge. 

Secondly, prior works split the same dataset into three sub-sets of different categories, which respectively serve as training, support, and query sets. Although they test on query sets with a new set of categories, it is still within the same semantic domain. In contrast, Tip-Adapter adapts the pre-trained CLIP into a totally new dataset for evaluation, which generalizes to a new domain and thus more challenging. Importantly, we test our models on full test sets, the same as conventional methods~\cite{he2016deep,dosovitskiy2021vit} trained by the full training set. Compared to existing works~\cite{vinyals2016matching,snell2017prototypical} on the small query sets, our effectiveness is verified by much more test images of new categories.

\section{Experiments}

\subsection{Training Settings}
\label{setting}
We conduct experiments for Tip-Adapter and Tip-Adapter-F on 11 widely-used image classification datasets: ImageNet \cite{deng2009imagenet}, StandfordCars \cite{krause20133d}, UCF101 \cite{soomro2012ucf101}, Caltech101 \cite{fei2004learning}, Flowers102 \cite{nilsback2008automated}, SUN397 \cite{xiao2010sun},  DTD \cite{cimpoi2014describing}, EuroSAT \cite{helber2019eurosat}, FGVCAircraft \cite{maji2013fine},  OxfordPets \cite{parkhi2012cats},  and Food101 \cite{bossard2014food}. For few-shot learning, we compare the performance of 1, 2, 4, 8, 16 few-shot training sets, and test on the full test sets. For the CLIP backbone, we utilize ResNet-50 \cite{he2016deep} as the visual encoder and a transformer \cite{dosovitskiy2021vit} as the textual encoder. We obtain the pre-trained weights of both encoders from~\cite{radford2021learning} and freeze them during training. We follow the data preprocessing protocol in CLIP~\cite{radford2021learning}, which is composed of random cropping, resizing, and random horizontal flip.
Other than the learnable prompts in CoOp, we follow CLIP to adopt prompt ensembling especially on ImageNet and use single handcrafted prompt on other 10 datasets. The Tip-Adapter-F is fine-tuned using batch size $256$, learning rate $0.001$, and the AdamW~\cite{kingma2014adam} optimizer with a cosine scheduler. We set 100-epoch training for EuroSAT dataset and only 20-epoch training for other 10 datasets.


Performance comparison is conducted between Zero-shot CLIP \cite{radford2021learning}, Linear-probe CLIP \cite{radford2021learning}, CoOp \cite{zhou2021coop} and CLIP-Adapter \cite{gao2021clip}. Therein, Zero-shot CLIP uses no extra training sample and conducts classification purely by pre-trained knowledge. Linear-probe CLIP trains an additional linear classifier after the weight-frozen CLIP on the few-shot training set. CoOp adopts learnable prompts for training, and we select its best-performing variant for comparison, that is, placing the class token at the end of the 16-token prompts without class-specific contexts. CLIP-Adapter appends a feature adapter \cite{houlsby2019parameter} to narrow the domain gap between the pre-trained features and downstream tasks. We also report the best-performing variant of CLIP-Adapter with only the learnable visual adapter. We report their official scores in the papers for fair comparison.

\begin{figure*}[t]
\begin{minipage}[t!]{0.39\linewidth}
    \centering
    \includegraphics[width=0.99\linewidth]{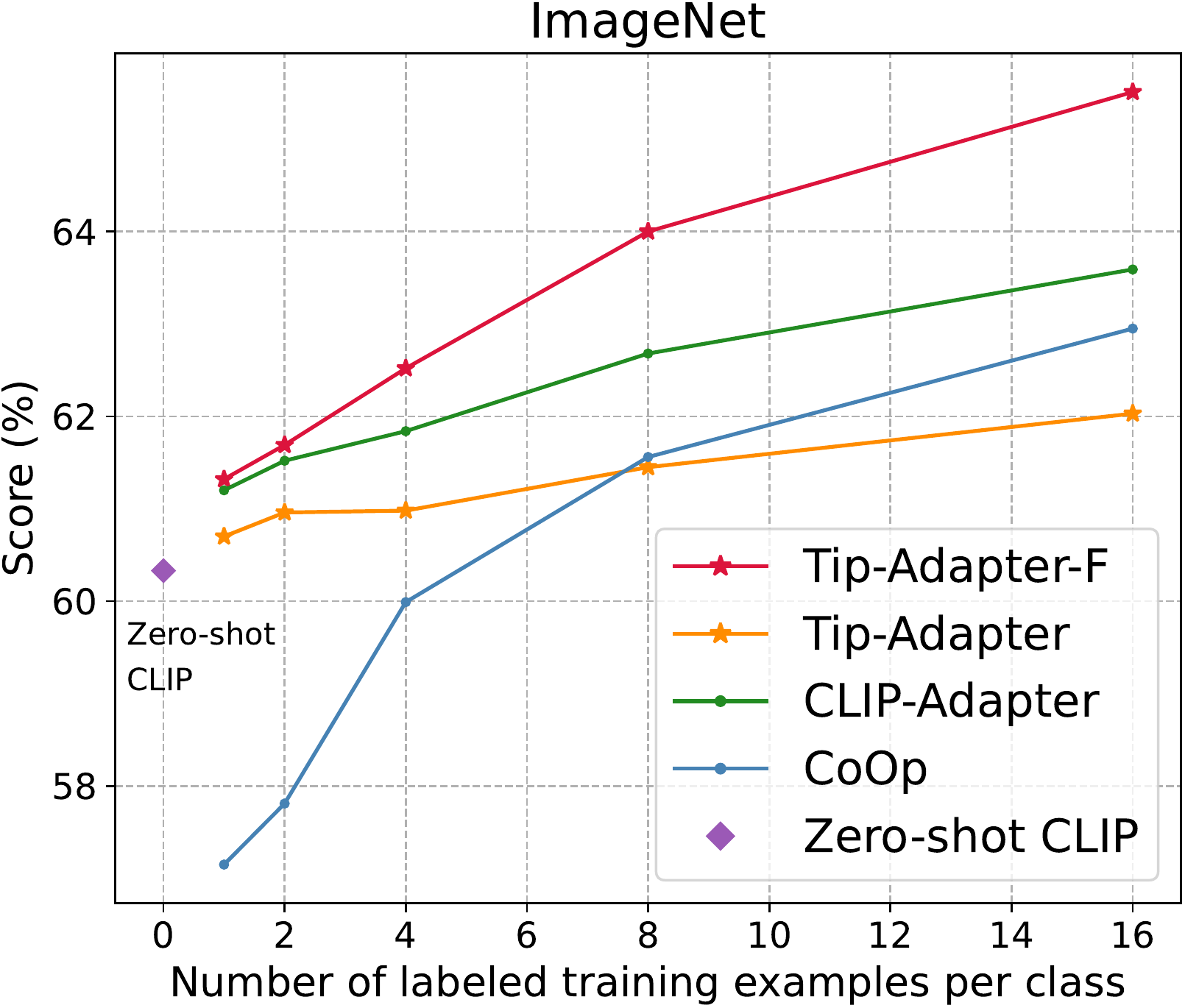}
    \figcaption{Few-shot classification accuracy of different models on ImageNet~\cite{deng2009imagenet}.}
    \label{imgfig}
\end{minipage}\qquad
\begin{minipage}[t!]{0.55\linewidth}
\begin{adjustbox}{width=\linewidth}
	\begin{tabular}{lcccccc}
	
	\toprule
	
		Few-shot Setup & 1  & 2  & 4  & 8 & 16 \\ \midrule
		
	    \multicolumn{6}{c}{\quad Zero-shot CLIP \cite{radford2021learning}: \ 60.33} \\  \cmidrule(lr){1-6}
	    Linear-probe CLIP \cite{radford2021learning}  &22.17 &31.90 &41.20 &49.52 &56.13\\
		CoOp \cite{zhou2021coop}  &57.15 &57.81 &59.99 &61.56 &62.95 \\
	    CLIP-Adapter \cite{gao2021clip}  &61.20  & 61.52 &  61.84 &  62.68 &  63.59 \\
	    \midrule
	    Tip-Adapter  &60.70   &60.96   &60.98   &61.45   &62.03 \\
	    Tip-Adapter-F  &\textbf{61.32}   &\textbf{61.69}   &\textbf{62.52} &\textbf{64.00}   &\textbf{65.51}  \\
	    &\color{blue}{+0.62} &\color{blue}{+0.73} &\color{blue}{+1.54} &\color{blue}{+2.55} &\color{blue}{+3.48} \\
	\bottomrule
	\end{tabular}
	\end{adjustbox}
	\tabcaption{Classification accuracy ($\%$) on ImageNet~\cite{deng2009imagenet} of different models with quantitative values. The last row in blue records the performance gain of Tip-Adapter-F brought by further fine-tuning over Tip-Adapter.}
	\label{table:acc}
\end{minipage}
\end{figure*}

\subsection{Comparison on ImageNet}

\paragraph{\textbf{Performance Analysis.}}
As shown in Figure~\ref{imgfig} and Table~\ref{table:acc}, both Tip-Adapter and Tip-Adapter-F show outstanding performance over other methods. Compared to Zero-shot CLIP, Tip-Adapter consistently surpasses it without any training. When the numbers of training samples are limited, Tip-Adapter greatly exceeds the Linear-probe CLIP by +38.53$\%$, +29.06$\%$ in 1-shot and 2-shot settings. With further fine-tuning, Tip-Adapter-F updates the keys in the cache model and achieves the best performance over all methods in all few-shot settings. The performance gain over Tip-Adapter becomes larger as the number of training samples increases, from 1-shot's +0.62$\%$ to 16-shot's +3.44$\%$. This indicates that the fine-tuning with more training samples enables the network to build a more powerful cache model. In Table~\ref{backbone}, we also implement different models with various visual encoders over ResNet~\cite{he2016deep} and ViT~\cite{dosovitskiy2021vit} backbones, where our Tip-Adapter-F still performs the best.

\begin{table}[t!]
\centering
\caption{Classification accuracy ($\%$) of different visual encoders on 16-shot ImageNet~\cite{deng2009imagenet}. ViT-B/32 and ViT-B/16 denote ViT-Base~\cite{dosovitskiy2021vit} with the patch size 32 $\times$ 32 and 16 $\times$ 16, respectively, and RN50$\times$16 denotes ResNet-50~\cite{he2016deep} with 16 times more computation~\cite{radford2021learning}. }
\begin{adjustbox}{width=0.75\linewidth}
	\begin{tabular}{lccccccc}
	\toprule
		Models &\ ResNet-50\ &\ ResNet-101\ &\ ViT-B/32\ & \ ViT-B/16\ &\ RN50$\times$16\\ \midrule
		Zero-shot CLIP~\cite{radford2021learning} &60.33 &62.53 &63.80 &68.73 &70.94\\
		CoOp~\cite{zhou2021coop} & 62.95 &66.60 & 66.85  &71.92 &-\\
		CLIP-Adapter~\cite{gao2021clip} & 63.59 & 65.39 & 66.19 &71.13 &-\\ \midrule
		Tip-Adapter & 62.03 & 64.78 & 65.61 & 70.75 &72.95\\
	    Tip-Adapter-F & \textbf{65.51} & \textbf{68.56} & \textbf{68.65} & \textbf{73.69} &\textbf{75.81}\\
	\bottomrule
	\end{tabular}
\end{adjustbox}
\label{backbone}
\end{table}

\paragraph{\textbf{Efficiency Comparison.}}
In Table~\ref{tab1}, we show the comparison of training time and inference speed for different models. CLIP-Adapter, Tip-Adapter and Tip-Adapter-F are able to cache the textual features from CLIP in the beginning and load them during training or inference, but CoOp adopts learnable prompts, which requires to calculate through the whole textual encoder online for every iteration. Linear-probe CLIP utilizes logistic regression~\cite{wright1995logistic}, so it cannot measure the training time by epochs and the inference speed on GPU. From the comparison, we observe that CoOp takes the most training time for learning prompts and has a +2.26$\%$ performance gain over Zero-shot CLIP. CLIP-Adapter significantly reduces the training time with better performance improvement of +3.26$\%$, but still needs 200-epoch training. Aided by the cache model, Tip-Adapter gains +1.70$\%$ improvement but requires no extra training time, which makes it a good trade-off between performance and efficiency. Tip-Adapter-F further reaches state-of-the-art accuracy with only $1/10$ of CLIP-Adapter and CoOp's training epochs, achieving the best of both worlds. As for inference speed and GPU memory consumption~\cite{10.5555/104279.104293}, our Tip-Adapter and Tip-adapter-F only produce marginal extra latency over Zero-shot CLIP and save much GPU memory compared to CoOp, which are quite efficient for applications.

    \begin{minipage}[h]{0.95\linewidth}
    \begin{minipage}[t]{0.32\linewidth}
    \includegraphics[width=\linewidth]{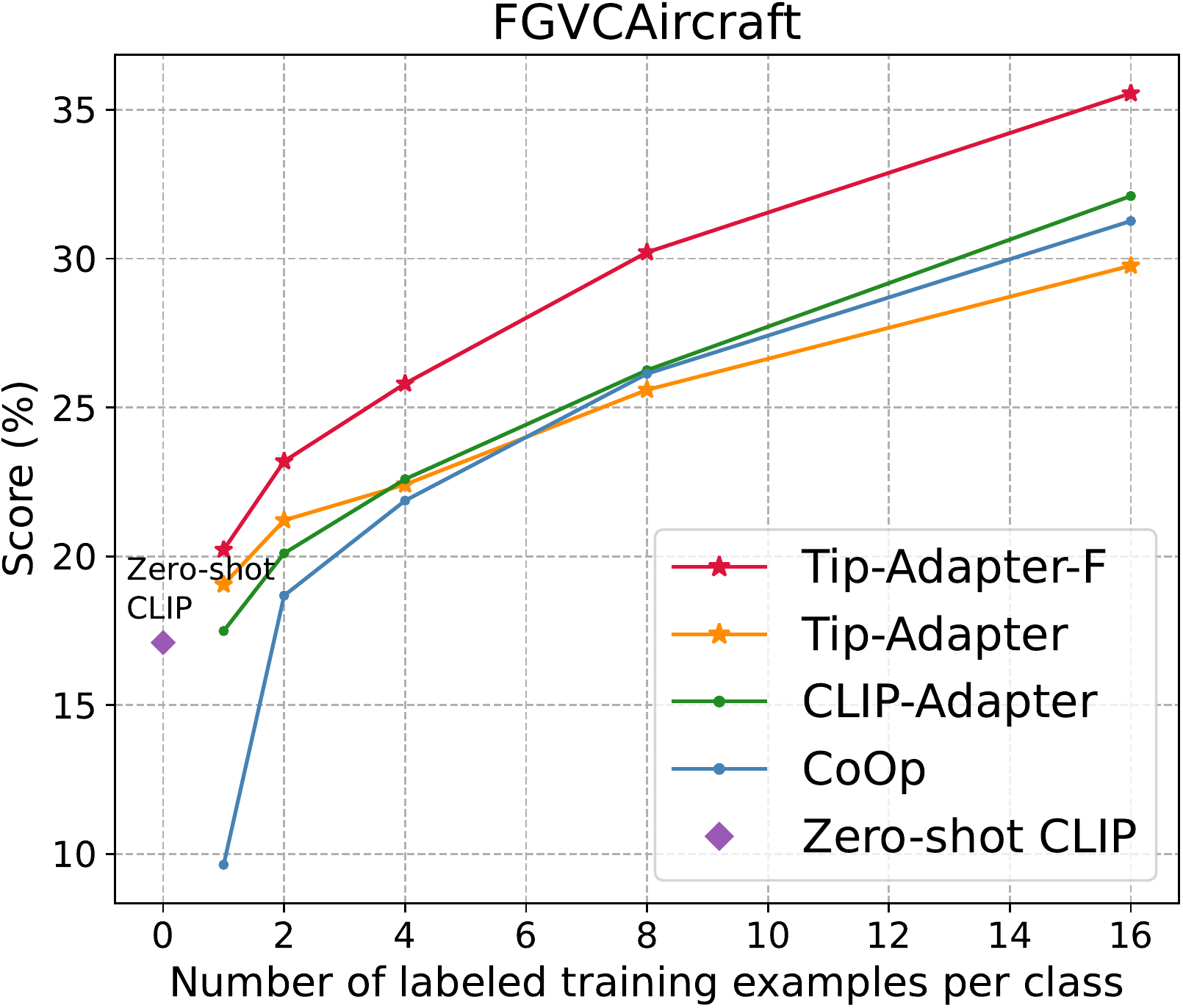}
    \end{minipage}
    \begin{minipage}[t]{0.32\linewidth}
    \includegraphics[width=\linewidth]{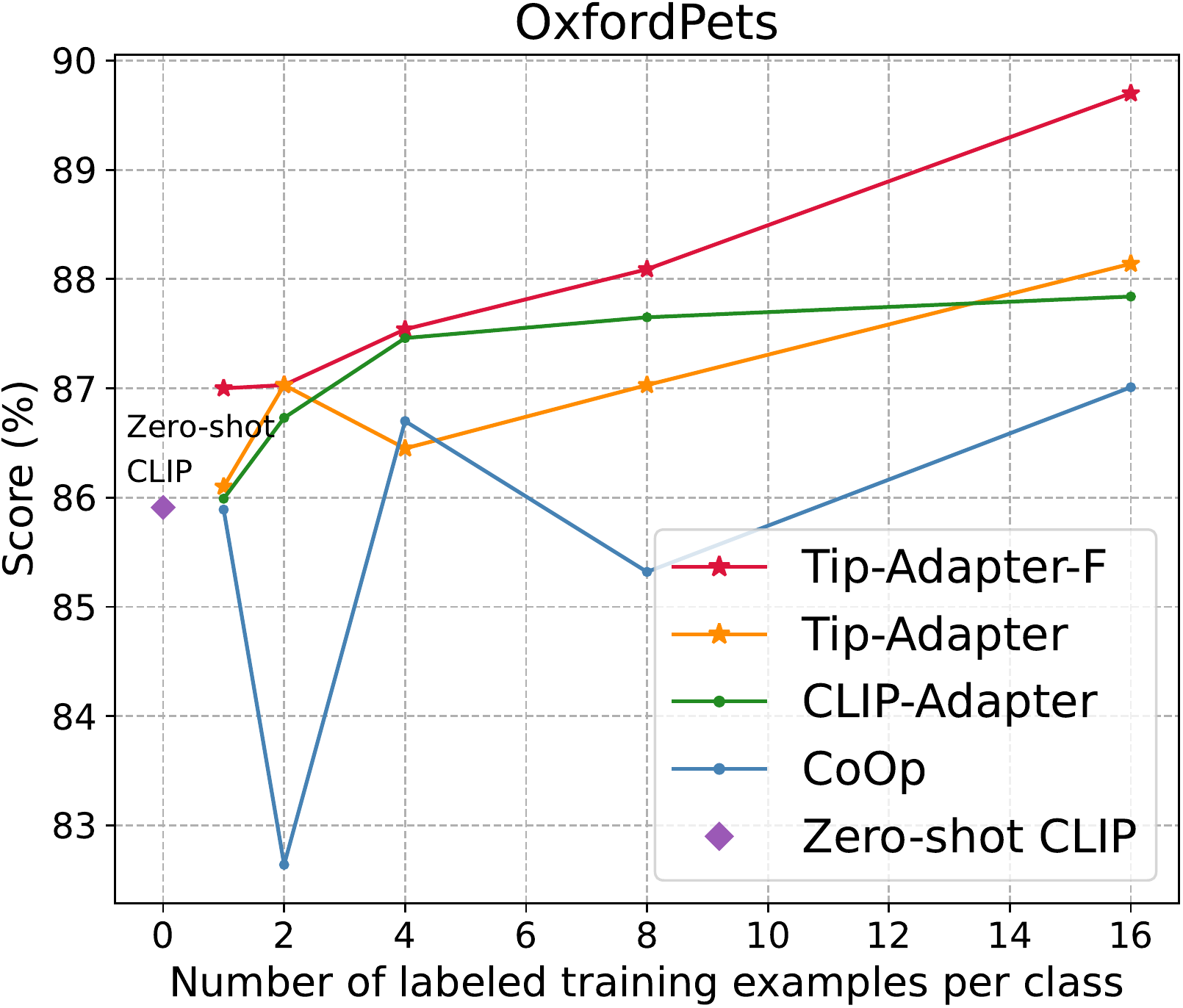}
    \end{minipage}
    \begin{minipage}[t]{0.32\linewidth}
    \includegraphics[width=\linewidth]{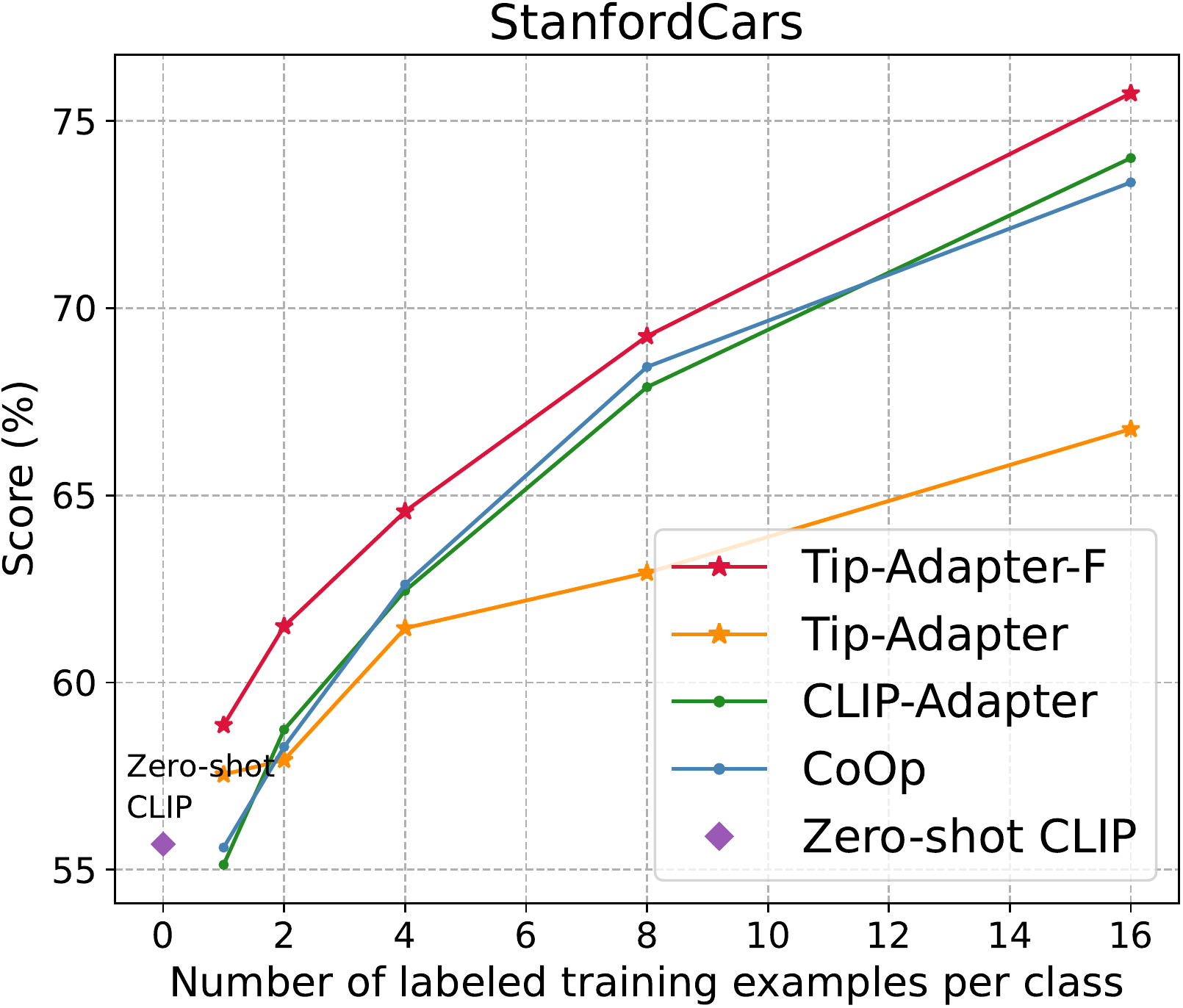}
    \end{minipage}
    \end{minipage}
    
    \begin{minipage}[h]{0.95\linewidth}
    \begin{minipage}[t]{0.32\linewidth}
    \includegraphics[width=\linewidth]{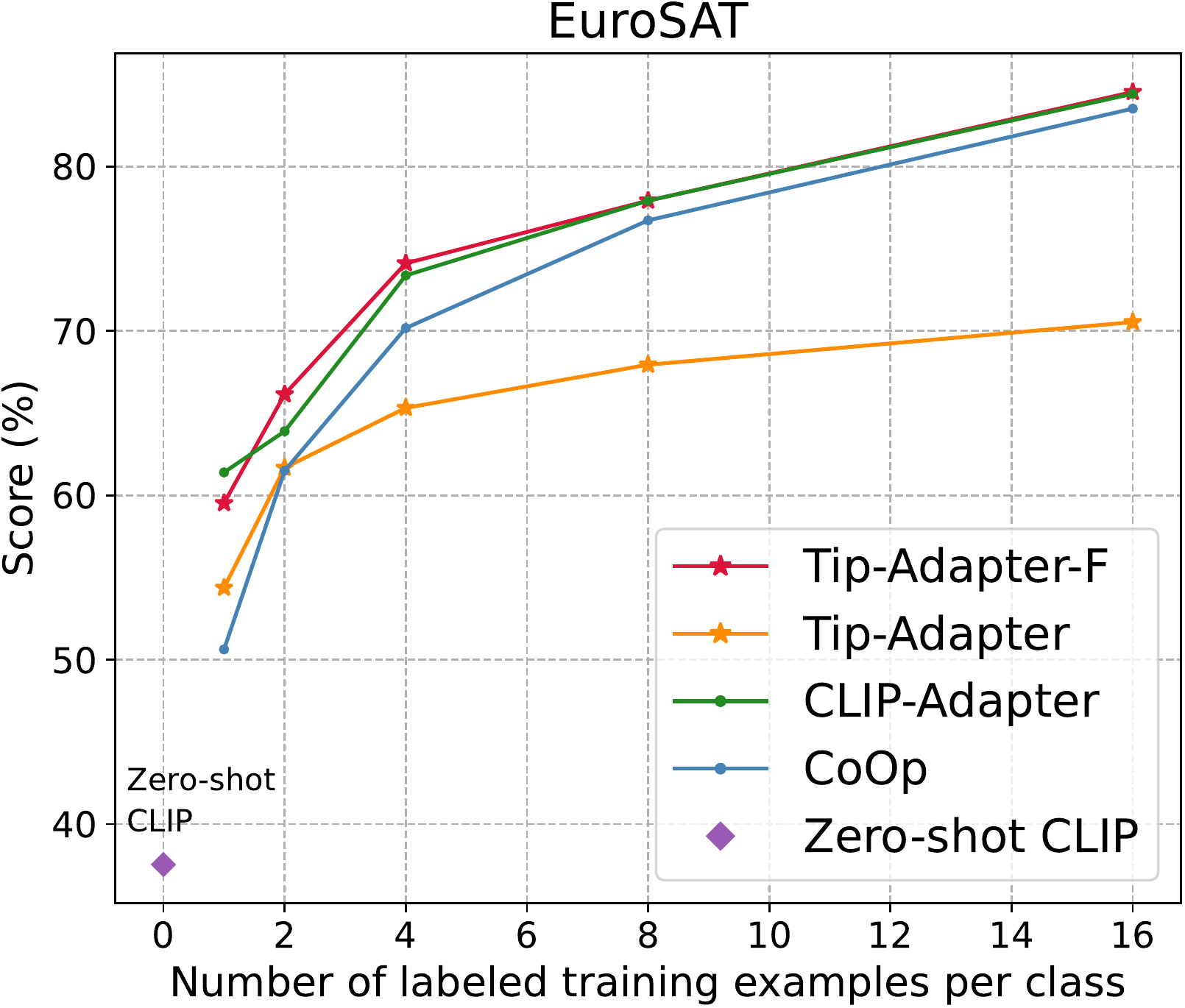}
    \end{minipage}
    \begin{minipage}[t]{0.32\linewidth}
    \includegraphics[width=\linewidth]{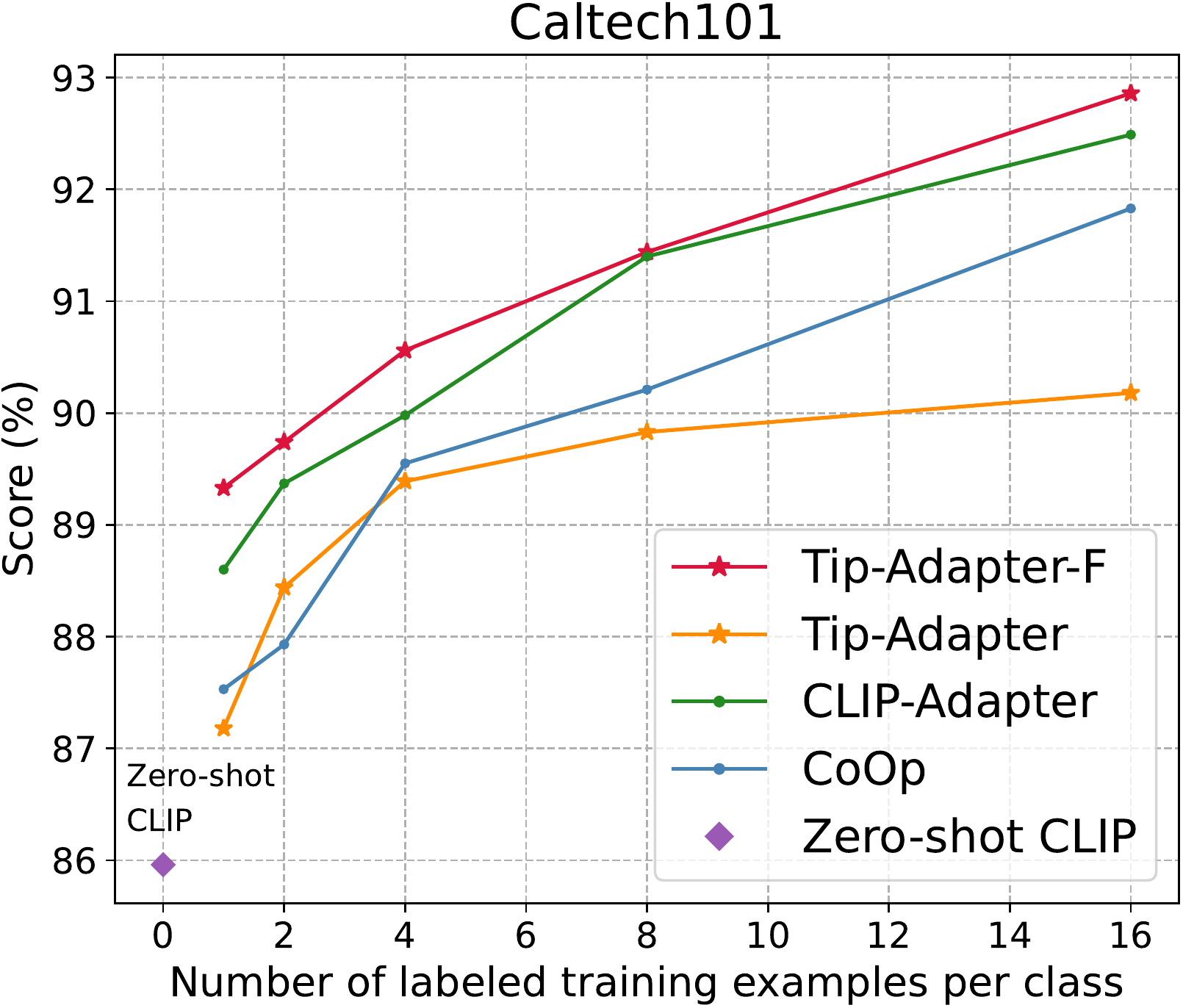}
    \end{minipage}
    \begin{minipage}[t]{0.32\linewidth}
    \includegraphics[width=\linewidth]{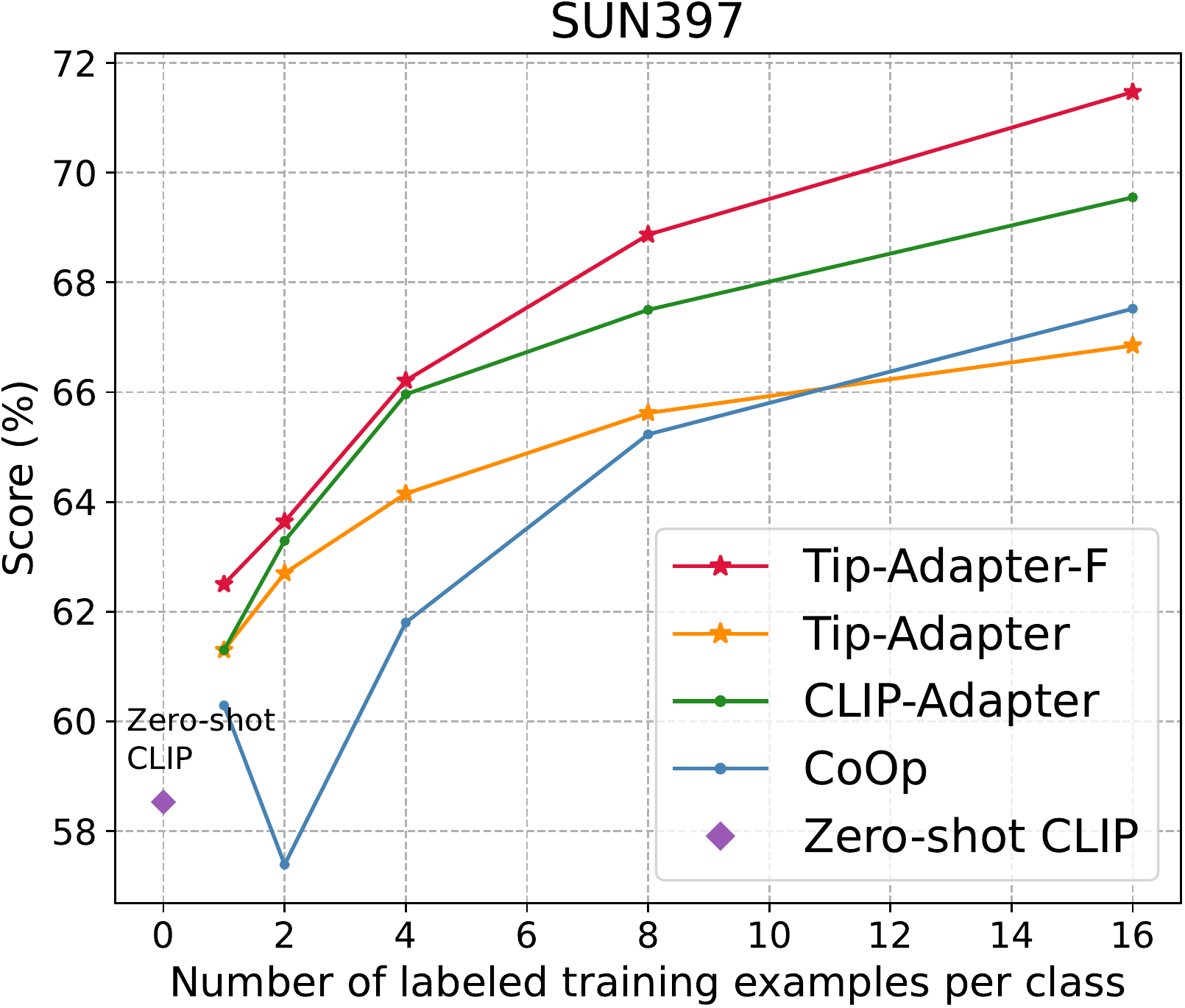}
    \end{minipage}
    \end{minipage}
    
    \begin{minipage}[h]{0.95\linewidth}
    \begin{minipage}[t]{0.32\linewidth}    
    \includegraphics[width=\linewidth]{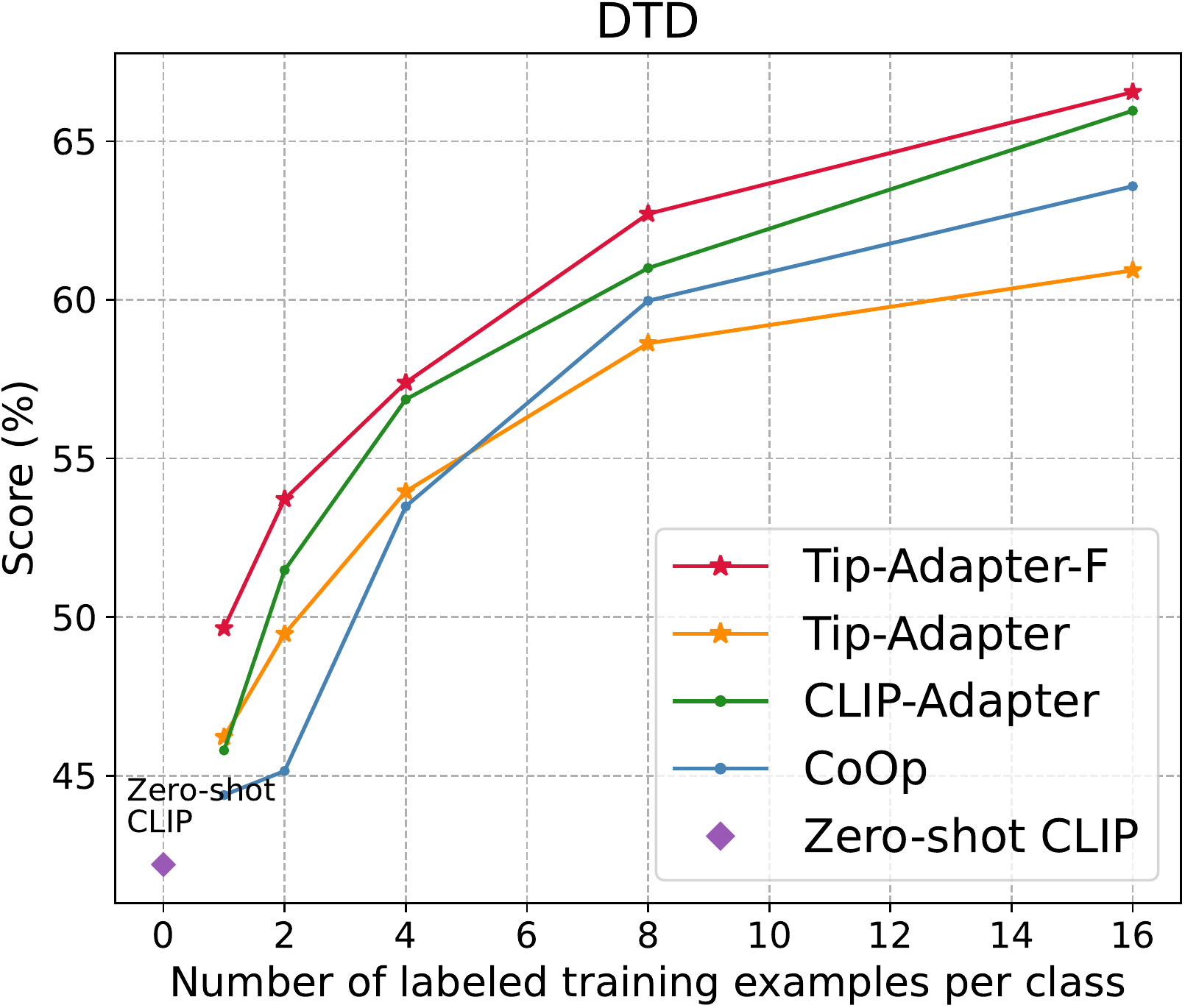}
    \end{minipage}
    \begin{minipage}[t]{0.32\linewidth}
    \includegraphics[width=\linewidth]{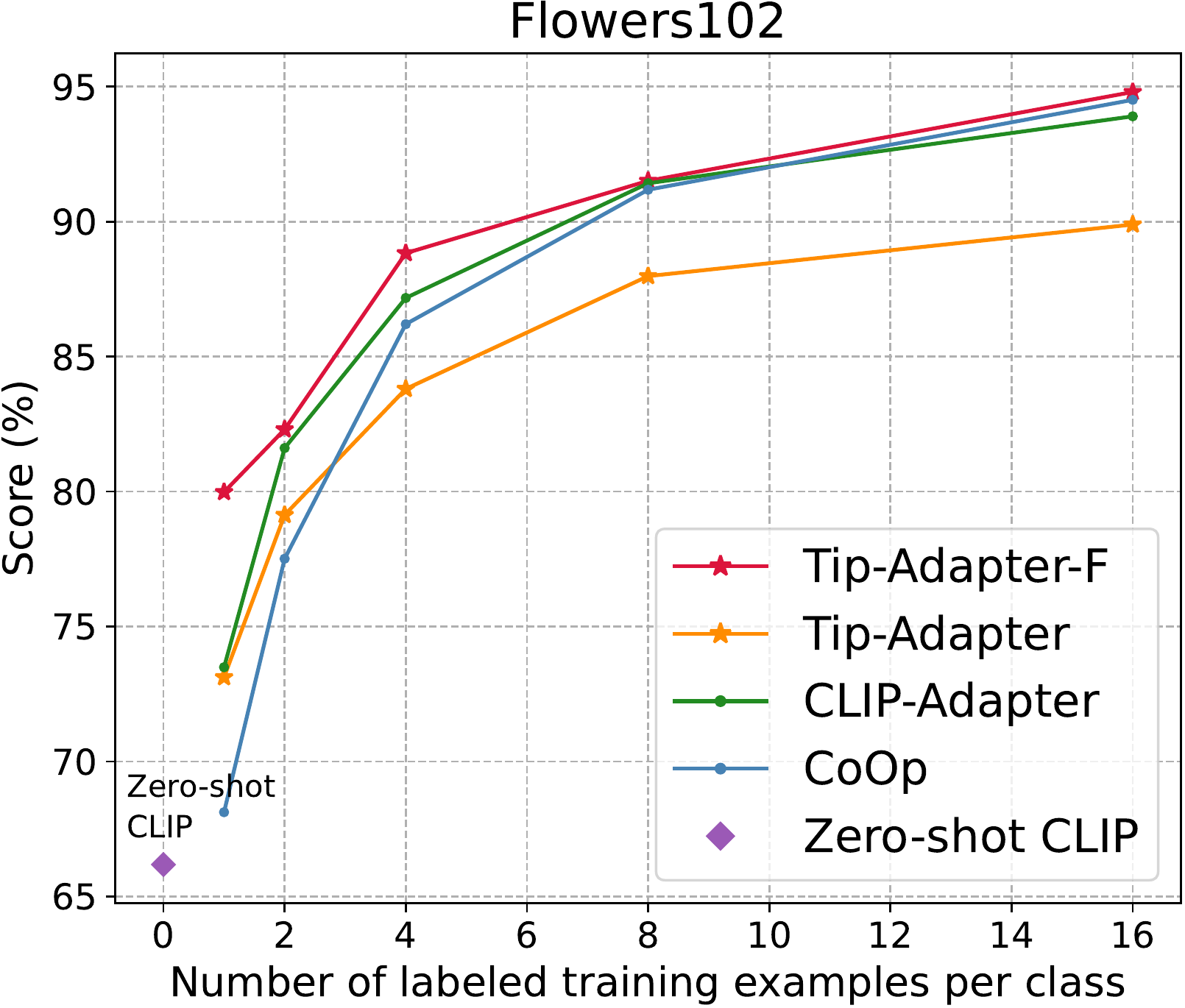}
    \end{minipage}
    \begin{minipage}[t]{0.32\linewidth}
    \includegraphics[width=\linewidth]{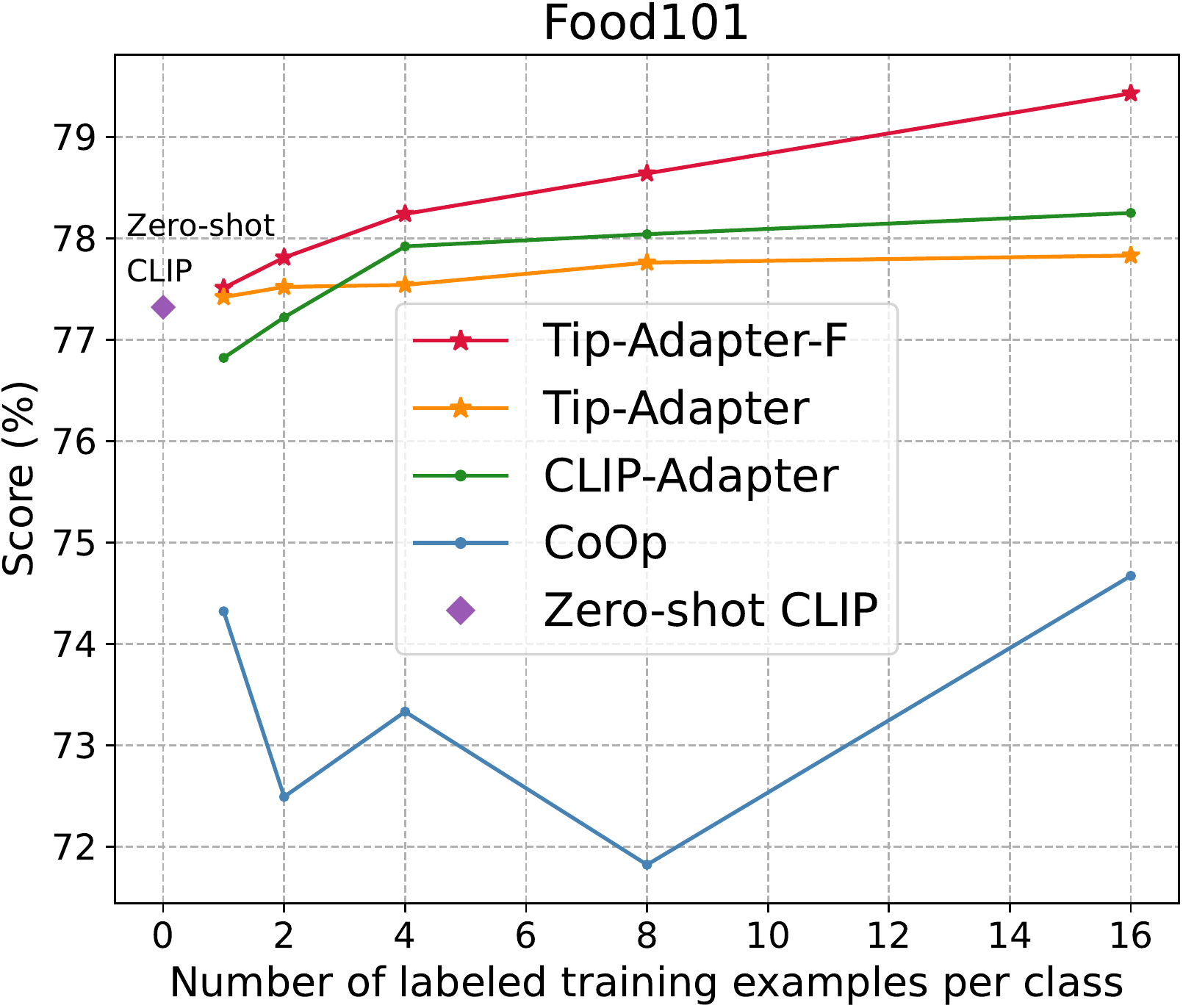}
    \end{minipage}
    \end{minipage}
    
    \begin{minipage}[h]{0.36\linewidth}
    \begin{minipage}[h]{\linewidth}
    \includegraphics[width=\linewidth]{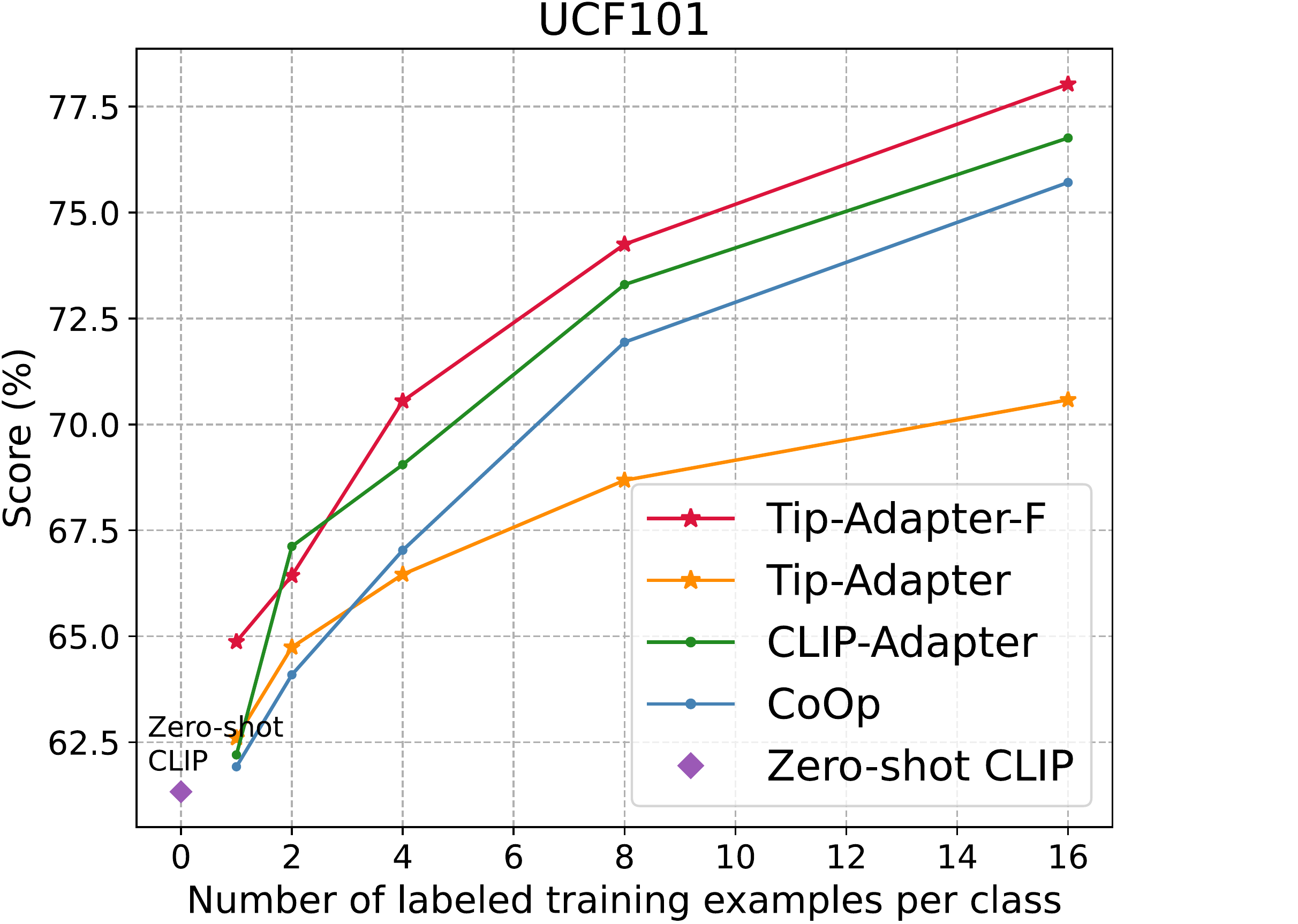}
    \end{minipage}
    \end{minipage}
    \begin{minipage}[h]{0.56\linewidth}
    \begin{minipage}[h]{\linewidth}
    \figcaption{Few-shot classification accuracy of different models on 10 datasets. Tip-Adapter largely improves Zero-shot CLIP without any training. Tip-Adapter-F consistently surpasses all compared methods by efficiently fine-tuning the cache model.}
    \label{fig:10_datasets}
    \end{minipage}
    \end{minipage}

\subsection{Performance on Other Datasets}

Figure~\ref{fig:10_datasets} shows the performance comparison on other 10 datasets listed in Section \ref{setting}. Our triaining-free Tip-Adapter significantly boosts the classification accuracy over Zero-shot CLIP and surpasses CoOp trained by 1 or 2 shots on most datasets. Although Tip-Adapter underperforms CoOp and CLIP-Adapter trained by more shots, Tip-Adapter-F with a fewer-epoch fine-tuning can eliminate the gap and further surpass all other models, achieving comprehensively leading performance. The consistent superiority of Tip-Adapter-F over 10 datasets fully demonstrates the effectiveness and generality of our proposed cache model.

\subsection{Ablation Studies}

In this section, we conduct several ablation studies about Tip-Adapter on ImageNet~\cite{deng2009imagenet}. All experiments adopt the 16-shot setting without training. 

\paragraph{\textbf{Residual Ratio $\alpha$.}}
The hyper-parameter $\alpha$ controls how much to combine newly adapted predictions from the cache model with pre-trained CLIP's, which can also be interpreted as weighing the visual and textual caches as in Eq.~\ref{logits_cache}. As formulated above, larger $\alpha$ denotes using more knowledge from the few-shot training set and less otherwise. We vary $\alpha$ from 0.0 to 5.0, and set the hyper-parameter $\beta$ as 5.5. When $\alpha$ equals 0.0, the model is equivalent to Zero-shot CLIP without using few-shot knowledge. From the top part of Table~\ref{alpha}, we observe that the classification accuracy is improving as $\alpha$ increases from 0.0 to 1.0, achieving the best 62.03$\%$ at 1.0. This indicates that the prior knowledge from CLIP and the few-shot knowledge from cache model are equally important.

\begin{table}[t]
\centering
\caption{Four ablation studies (\%) of Tip-Adapter on ImageNet~\cite{deng2009imagenet}, from top to bottom: residual ratio $\alpha$, sharpness ratio $\beta$, the size of cache model, and the performance given more shots while fixing cache size 16.
}
\begin{adjustbox}{width=0.55\linewidth}
	\begin{tabular}{c|cccccc}
	\toprule
	   
		\multicolumn{7}{c}{Ablation Studies on Tip-Adapter} \\ 
		\midrule
		\multirow{2}{*}{\shortstack{Residual Ratio\\$\alpha$}}
		  &0.0 &0.5 &\textbf{1.0} &\ \ 2.0\  &\ 3.0\ \ &\ 4.0 \\  
        \cmidrule(lr){2-7}
		 &\ 60.33  & 61.44  &\textbf{62.03} &\ \ 61.41 & \ 60.36 &\ 59.14\\ 
        \midrule
		
	    \multirow{2}{*}{\shortstack{Sharpness Ratio\ \  \\$\beta$}} &1.5 &3.5 &\textbf{5.5} &7.5 &9.5 &11.5 \\
	     \cmidrule(lr){2-7}
	    &\ 61.82  &61.91 &\textbf{62.03} &\ \ 61.76 &\ 61.62 &\ 61.40\\
	    
	   \midrule
	   \multirow{2}{*}{\shortstack{Cache Size \ } }&0 &1 &2 &4 &8 &\textbf{16} \\
	     \cmidrule(lr){2-7}
	    &\ 60.33  &61.45 &61.71 &61.79 &61.83 &\textbf{62.03}\\
	    
	    \midrule
	    \midrule
	   \multirow{3}{*}{\shortstack{More Shots\\than 16}} &\multicolumn{2}{l}{\ Shot Setup} &16 &32 &64 &128 \\
	     \cmidrule(lr){2-7}
	    &\multicolumn{2}{l}{\ Tip-Adapter} &62.03 &62.51 &62.88 &63.15\\
	     &\multicolumn{2}{l}{\ Tip-Adapter-F} &65.47 &66.58 &67.96 &69.74\\
	\bottomrule
	\end{tabular}
\end{adjustbox}
\label{alpha}
\end{table}

\paragraph{\textbf{Sharpness Ratio $\beta$.}}
In Eq.~\eqref{beta}, $\beta$ in the activation function $\varphi$ controls the sharpness of the affinities. When $\beta$ is large, only the most similar training samples to the test image in the embedding space have the large influences to the prediction and vice versa. In the second part of Table~\ref{alpha} with the $\alpha$ as 1.0, we observe that the variation of $\beta$ has a limited impact and a moderate 5.5 for $\beta$ leads to the best-performing Tip-Adapter.

\paragraph{\textbf{Size of the Cache Model.}}
\label{size}
We explore the influence of the size for cache model in Tip-Adapter. Given 16-shot training set, rather than caching all 16 samples per category, we construct the cache whose size is more than 0 but less than 16. Taking 8 as an example, we randomly divide 16 samples into 8 uniform groups and obtain 8 prototypes by averaging features of the 2 samples in each group. Considering such random division of samples might influence the performance, we experiment 5 times and report the average scores. The results from the third part of Table~\ref{alpha} illustrate that, the more samples we cache to preserve more few-shot knowledge, the higher accuracy Tip-Adapter can achieve.


\paragraph{\textbf{Scaling up to More Shots.}}
Given more than 16 shots, we explore a way to still constrain the cache size as 16 and avoid the potential burden for both memory and computation. Taking 64 shots as an example, following the division strategy in the above paragraph, we obtain 16 prototypes from 4 groups to construct the cache model. The final part of Table~\ref{alpha} indicates that even if the cache size is restrained to 16, Tip-Adapter can well capture the knowledge from 32, 64, and 128 training samples per category. Also, the performance boost gradually slows down when more samples are provided, which implies a possible limit of cache size 16 without training. However, Tip-Adapter-F can break such limit by fine-tuning the keys and achieve better performance by more shots for training.

\begin{figure*}[t!]
\begin{minipage}{0.45\linewidth}
    \label{fig:ablation}
    \includegraphics[width=\linewidth]{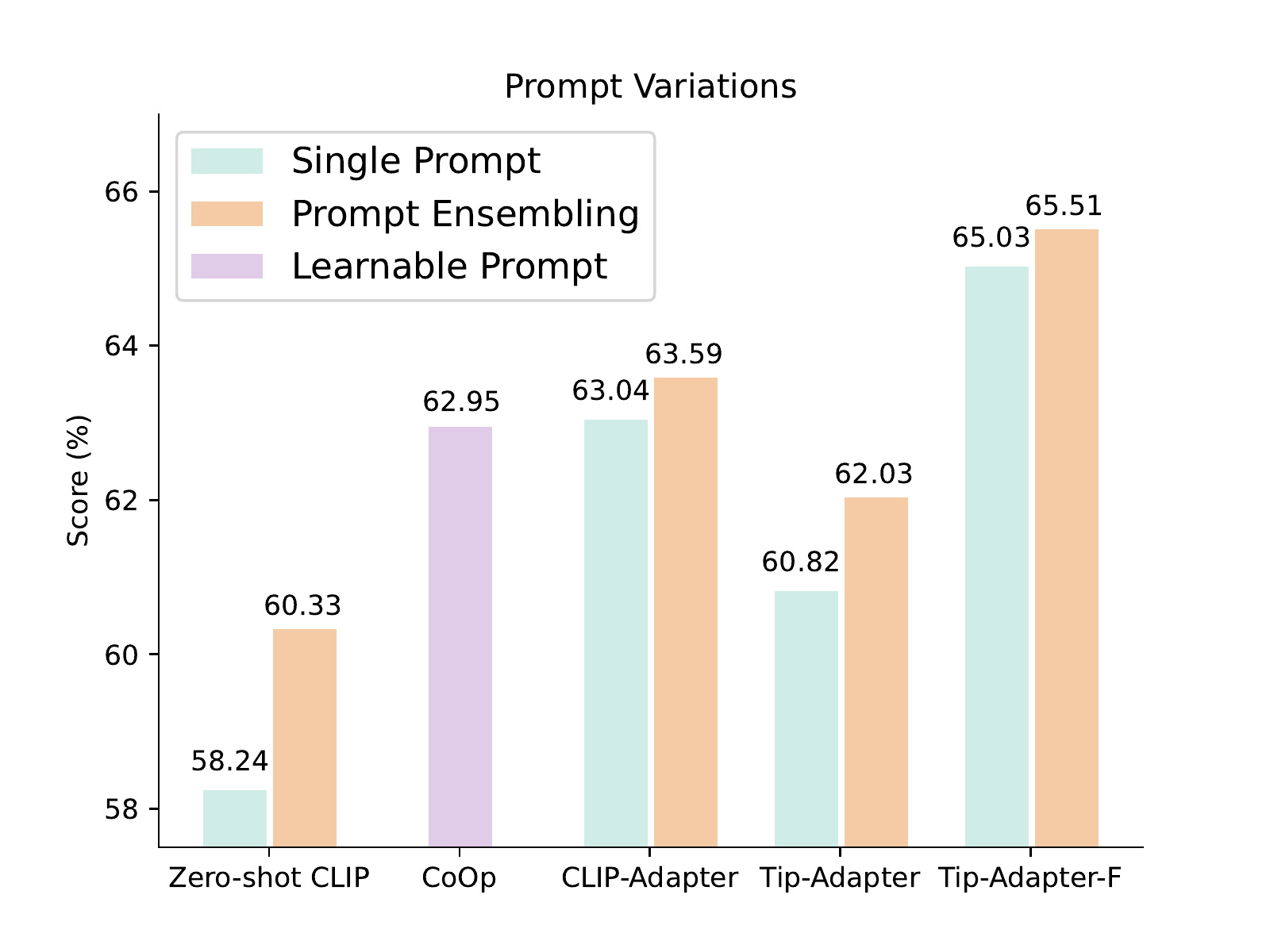}
    \figcaption{Classification performance with different prompt designs: single prompt (Cyan), prompt ensembling (Orange) and learnable prompt (Purple).}
     \label{prompt}
\end{minipage}\quad
\begin{minipage}{0.49\linewidth}
\centering
\begin{adjustbox}{width=\linewidth}
\begin{tabular}{lcccc}
\toprule
\multirow{3}{*}{Datasets} & \textbf{Source} &\multicolumn{2}{c}{\textbf{Target}} \\
\cmidrule(lr){2-2} \cmidrule(lr){3-4} 
& ImageNet  & -V2 & -Sketch  \\
&~\cite{deng2009imagenet} &~\cite{recht2019imagenet} &~\cite{hendrycks2021natural}\\ \midrule
Zero-Shot CLIP~\cite{radford2021learning}  & 60.33  & 53.27 & 35.44\\
Linear Probe CLIP~\cite{radford2021learning}  & 56.13  & 45.61 & 19.13\\
CoOp~\cite{zhou2021coop} & 62.95  & 54.58 & 31.04  \\
CLIP-Adapter~\cite{gao2021clip} &  {63.59}  &  {55.69} &  {35.68} \\
\midrule
Tip-Adapter &  {62.03}  &  {54.60} &  {35.90} \\
Tip-Adapter-F & \textbf{65.51}  & \textbf{57.11} & \textbf{36.00} \\
&\color{blue}{+3.48} &\color{blue}{+2.51} &\color{blue}{+0.10} \\
\bottomrule
\end{tabular}
	\end{adjustbox}
	\tabcaption{The robustness (\%) to distribution shift of different methods. The last row in blue records the performance gain of Tip-Adapter-F brought by further fine-tuning over Tip-Adapter..}
\label{table:robust}
\end{minipage}
	
\end{figure*}

\paragraph{\textbf{Prompt Design.}}
We utilize prompt ensembling of 7 templates from~\cite{radford2021learning} for Zero-shot CLIP, CLIP-Adapter, and Tip-Adapter as default. In Figure~\ref{prompt}, we test them only using a single prompt, ``a photo of a [CLASS].'', and observe slightly worse performance. The accuracy drops are smaller for Tip-Adapter-F and CLIP-Adapter, but larger for Tip-Adapter and Zero-shot CLIP, which indicates the better-performing models are less affected by the prompt variations.

\subsection{Distribution Shift}
We evaluate the out-of-distribution ability of our proposed Tip-Adapter and Tip-Adapter-F by learning from one dataset but testing on another. We set ImageNet~\cite{deng2009imagenet} as the source dataset providing 16-shot training set, and adopt two target datasets for testing: ImageNetV2~\cite{recht2019imagenet} and ImageNet-Sketch~\cite{hendrycks2021natural}, which contain compatible categories to ImageNet but with semantic gaps.
As shown in Table~\ref{table:robust}, Tip-Adapter without training exerts superior robustness to distribution shift, which surpasses CoOp~\cite{zhou2021coop} on ImageNet-V2 and CLIP-Adapter~\cite{gao2021clip} on ImageNet-Sketch. This indicates the cache model is more advantageous to out-of-distribution evaluation, whose training-free construction alleviates the risk of over-fitting on the source dataset.
Further, Tip-Adapter-F achieves the best of both worlds: the strong out-of-distribution performance brought by cache model and the leading in-distribution ability by fine-tuning.


\section{Visualization}
\begin{figure*}[t!]
  \centering
    \includegraphics[width=0.88\textwidth]{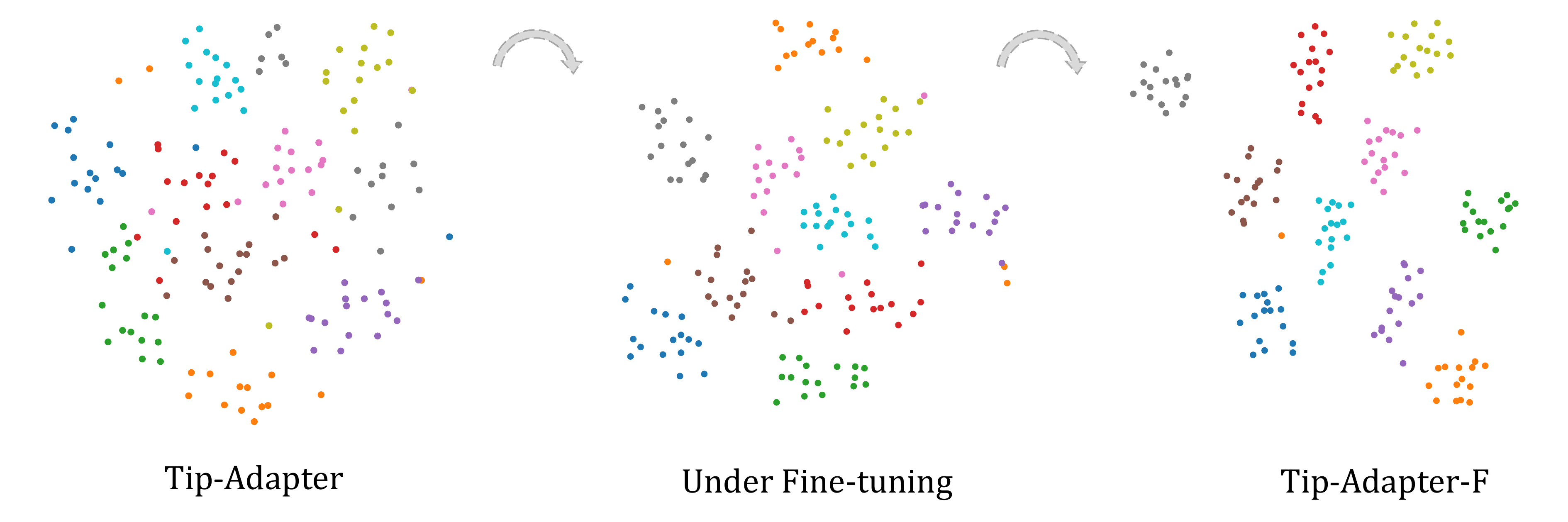}
   \caption{t-SNE visualization of $\mathbf{F}_\mathrm{train}$ in Tip-Adapter. Dots in different colors stand for embeddings of different categories. From left to right, three distributions indicate the variation of keys in cache model during fine-tuning.}
    \label{tsne}
\end{figure*}

To better show the variation of cache model during fine-tuning, we utilize t-SNE ~\cite{radford2021learning} to visualize the keys $\mathbf{F}_\mathrm{train}$ in Figure~\ref{tsne}. The dots in different colors denote 10 categories of 16-shot ImageNet~\cite{deng2009imagenet}, and their relative distances reflect the high-dimensional distributions of category embeddings.
From left to right, the three sub-figures represent the training-free Tip-Adapter, Tip-Adapter during fine-tuning and the final Tip-Adapter-F, respectively. It could be observed that before training, the distribution has shown good discrimination thanks to the properly designed cache model construction. During fine-tuning, embeddings of the same category gradually converges together and different clusters become more contrastive and separate, contributing to stronger classification capability.

\section{Conclusions} 
\label{sec:conclusion}
We propose Tip-Adapter, a non-parametric adaption method of CLIP, which acquires the adapter by a cache model constructed from the few-shot training set. In this way, the few-shot knowledge is retrieved from the cache model and incorporated with CLIP's pre-trained knowledge in a training-free manner. On top of that, Tip-Adapter can be further enhanced by fine-tuning the cached keys for just a few epochs, named Tip-Adapter-F, which achieves state-of-the-art performance among existing methods. Considering limitations, although it is marginal, Tip-Adapter-F still requires 20-epoch fine-tuning on ImageNet to learn the best-performing cache model. Our future work will focus on exploring new training-free methods for CLIP to fully unleash its power for visual representation.

\paragraph{{\rm {\bf Acknowledgement.}}}
This work is supported in part by Centre for Perceptual and Interactive Intelligence Limited, in part by the General Research Fund through the Research Grants Council of Hong Kong under Grants (Nos. 14204021, 14207319), in part by CUHK Strategic Fund, and in part by the Shanghai Committee of Science and Technology (Grant No. 21DZ1100100).

\appendix

\paragraph{{\rm {\bf Appendix}}}

\section{Fine-tuning Settings}
Compared to Tip-Adapter without training, Tip-Adapter-F fine-tunes the keys $\mathbf{F}_\mathrm{train}$ in the cache model, but freezes values $\mathbf{L}_\mathrm{train}$, CLIP's~\cite{radford2021learning} visual encoder and textual encoder. Here, we explore whether other modules in Tip-Adapter could be fine-tuned for performance improvement. In Table~\ref{fine-tune}, we conduct 7 fine-tuning experiments for unfreezing different modules of Tip-Adapter. Note that we set the learning rates of two CLIP's encoders as 1/1000 of the $\mathbf{F}_\mathrm{train}$ and $\mathbf{L}_\mathrm{train}$'s for training stability, and train every settings for 20 epochs on ImageNet~\cite{deng2009imagenet} with 16-shot training set. As shown, the first two rows denote the performance for Tip-Adapter's 62.03$\%$ and Tip-Adapter-F's 65.51$\%$. The third row by fine-tuning the cached values $\mathbf{L}_\mathrm{train}$ decreases the performance to 60.90$\%$, and fine-tuning all cache model even leads to collapse during training, which accords with our assumption that the one-hot ground-truth labels shall not be updated to preserve the few-shot knowledge. Furthermore, we experiment to fix all parameters in the cache model and fine-tune the pre-trained CLIP's weights. If the visual encoder or textual encoder is independently tuned, the performance could be improved to 62.84$\%$ and 63.15$\%$, respectively, but when both encoders are jointly fine-tuned, the classification accuracy would significantly drop to 51.22$\%$. This is because of the severe over-fitting for such a huge-parameter model learning from the few-shot training set. Compared to unfreezing CLIP's encoders, only fine-tuning $\mathbf{F}_\mathrm{train}$ brings larger performance improvement but less time consumption, which fully demonstrates the superiority of our Tip-Adapter-F.

\begin{table}[h!]
\centering
\caption{Fine-tuning different modules for Tip-Adapter. `\Checkmark' denotes fine-tuning and the symbol `-' denotes freezing. `Vis.' and `Tex.' stand for visual encoder and textual encoder of CLIP. The accuracy ($\%$) and training time are tested on 16-shot ImageNet~\cite{deng2009imagenet} and a single NVIDIA GeForce RTX 3090 GPU.}
\begin{adjustbox}{width=0.5\linewidth}
\begin{tabular}{ccccccc}
	\toprule
		 Vis.\ \ \ & Tex.\ \ \
		 &\ $\mathbf{F}_\mathrm{train}$\ &\ $\mathbf{L}_\mathrm{train}$\ &\ \ Accuracy &\ Time\\ \midrule
		- &- &- &- &62.03 &\textbf{0}\\
		- &- &\Checkmark &- &\textbf{65.51} &5min\\
		- & - & - & \Checkmark &60.90 &5min\\
		- & - & \Checkmark & \Checkmark &Collapsed &-\\
		\Checkmark & - & - & - &62.84 &8min\\
		- & \Checkmark & - & - & 63.15 &1h 20min\\
	    \Checkmark & \Checkmark & - & - & 51.22 &1h 27min\\
	\bottomrule
	\end{tabular}
\end{adjustbox}
\label{fine-tune}
\end{table}

\section{Performance Gain without Training}

\begin{figure*}[h]
  \centering
    \includegraphics[width=0.75\textwidth]{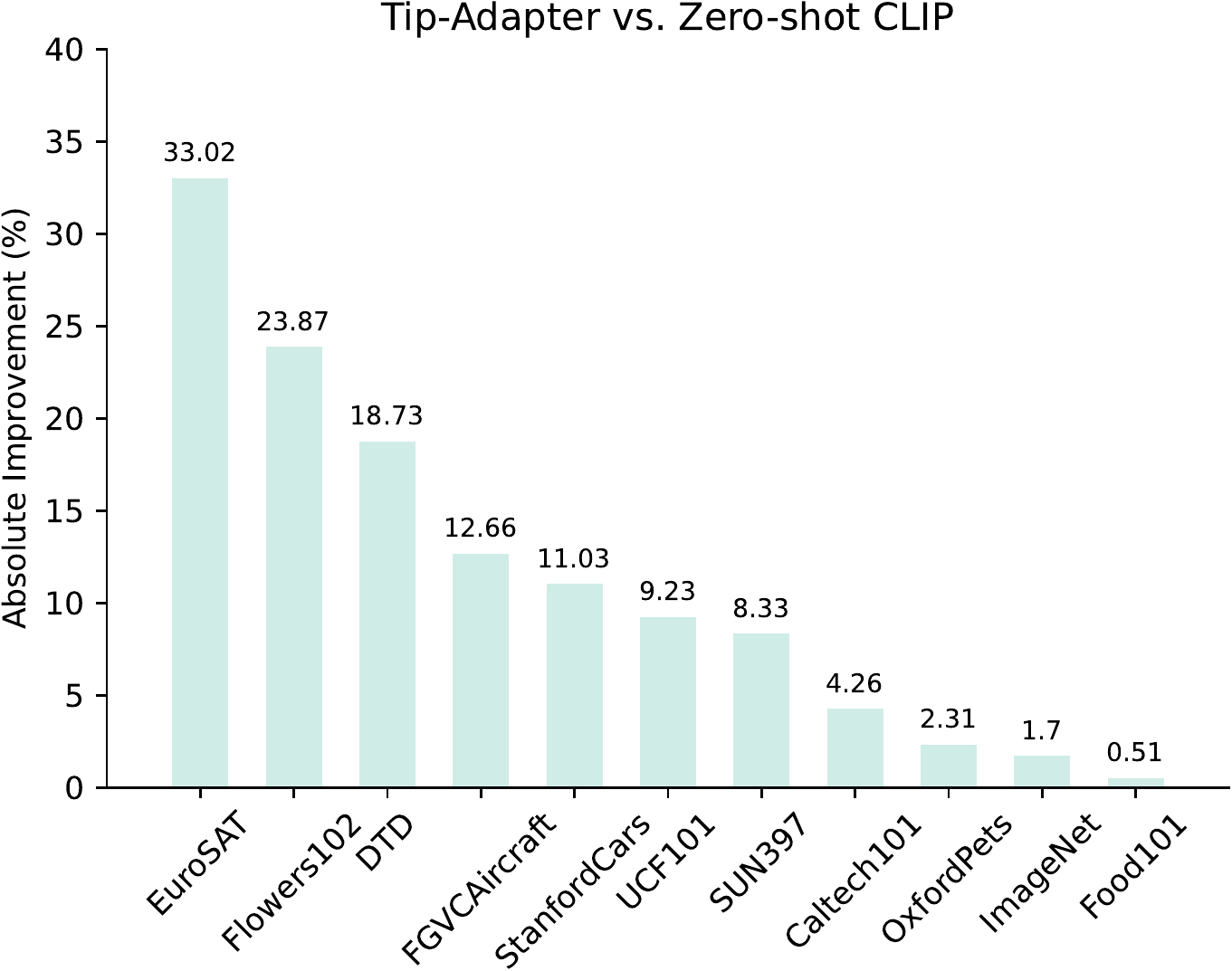}
   \caption{Performance gain contributed from the proposed training-free cache model, which is constructed by the 16-shot training set on 11 classification datasets.}
    \label{gain}
\end{figure*}

In Figure~\ref{gain}, we show the absolute accuracy improvement brought by Tip-Adapter over Zero-shot CLIP~\cite{radford2021learning} on 11 classification datasets under 16-shot settings. Without any training, Tip-Adapter greatly boosts Zero-shot CLIP on EuroSAT by 33.02$\%$ and Fowers102 by 23.87$\%$. Now that the CLIP is pre-trained on large-scale web-collected image-text pairs for daily scenarios, when the domain gap between downstream dataset and the pre-trained data is larger, the performance gain by Tip-Adapter would be normally higher. Taking EuroSAT and DTD as examples, they respectively contain land cover and detailed texture pictures with distinctive semantics, which thus require more few-shot knowledge memorized in the cache model to update the pre-trained CLIP's knowledge for better performance.

\section{Compared to Fully-trained Methods}
Although our Tip-Adapter and Tip-Adapter-F are based on the few-shot training sets, they are evaluated by the full test sets, the same as conventional methods~\cite{he2016deep,dosovitskiy2021vit} trained by full training sets. In Table~\ref{conven}, we compare the learnable parameters and training settings between ours and the series of ResNet~\cite{he2016deep} and DeiT~\cite{touvron2021training}. We adopt ViT-Large~\cite{dosovitskiy2021vit} as the visual backbone of Tip-Adapter and Tip-Adapter-F. As shown, only by 16-shot training set, Tip-Adapter without parameters or training outperforms ResNet-50 and DeiT-T by +1.9\% and +3.9\%, respectively. Tip-Adapter-F further achieves higher performance by the efficient fine-tuning of 6 minutes. This demonstrates the superiority of our approach in low-data and resource-limited regimes.

\begin{table}[h!]
\centering
\caption{Comparison between Tip-Adapter, Tip-Adapter-F and conventional methods trained by full training set on ImageNet~\cite{deng2009imagenet}.
The training time is tested on a single NVIDIA GeForce RTX 3090 GPU.}
\begin{adjustbox}{width=0.7\linewidth}
\begin{tabular}{ccccc}
	\toprule
		 Method &Acc. (\%)\ &Param. (M)\ \ &Train. Set\ \ &Train. Time\ \ \\ \cmidrule(lr){1-1} \cmidrule(lr){2-5}
		ResNet-50~\cite{he2016deep} &74.2 &25.6  &full set &\textgreater 1 day\\
	    ResNet-101~\cite{he2016deep} &77.4 &44.5  &full set &\textgreater 1 day\\
	    \cmidrule(lr){1-5}
	    DeiT-T~\cite{touvron2021training} &72.2 &6.0  &full set &\textgreater 1 day\\
	    DeiT-S~\cite{touvron2021training} &\textbf{79.9} &22.1  &full set &\textgreater 1 day\\
	    \cmidrule(lr){1-5}
	    Tip-Adapter &76.1 &\textbf{0}  &\textbf{16-shot} &\textbf{0}\\
	    Tip-Adapter-F &79.4 &6.2  &\textbf{16-shot} &6 min\\
	\bottomrule
	\end{tabular}
\end{adjustbox}
\label{conven}
\end{table}

\clearpage
%
%
\bibliographystyle{splncs04}
\bibliography{egbib}
\end{document}